\documentclass[sn-apa]{sn-jnl}

\usepackage{graphicx}
\usepackage{rotating} %
\usepackage{multirow}%
\usepackage{amsmath,amssymb,amsfonts}%
\usepackage{amsthm}%
\usepackage{mathrsfs}%
\usepackage[title]{appendix}%
\usepackage{xcolor}%
\usepackage{textcomp}%
\usepackage{manyfoot}%
\usepackage{booktabs}%
\usepackage{algorithm}%
\usepackage{algorithmicx}%
\usepackage{algpseudocode}%
\usepackage{listings}%
\usepackage{subcaption}
\usepackage{amsmath}
\usepackage{algorithm}
\usepackage{algpseudocode}
\usepackage{float}

\usepackage{verbatim}

\theoremstyle{thmstyleone}%
\theoremstyle{thmstyletwo}%

\newcommand{\celltwo}[2]{%
\shortstack[c]{%
#1\\[-0.15em]
{\footnotesize (#2)}%
}%
}
\theoremstyle{thmstylethree}%

\newcount\Comments  
\usepackage{color}
\definecolor{darkgreen}{rgb}{0,0.5,0}
\definecolor{purple}{rgb}{1,0,1}
\newcommand{\kibitz}[2]{\ifnum\Comments=1\textcolor{#1}{#2}\fi}

\begin{document}

\title[]{
 Robust State-Conditional Feature-Weighted Jump Models for Temporal Clustering
}
\author{
\centering
Federico P. Cortese$^{1,3}$ and Alessio Farcomeni$^{2}$\\[0.2cm]
\small $^{1}$University of Milan, Department of Economics, Management, and Quantitative Methods\\[0.1cm]
\small $^{2}$University of Rome ``Tor Vergata'', Department of Enterprise Engineering\\[0.1cm]
\small $^{3}$National Research Council of Italy, Institute for Applied Mathematics and Information Technologies, Milan
}

\abstract{
We propose a robust feature-weighted jump model for time-dependent clustering. 
A penalty is used to encourage smoothness of transitions over time, while robustness is achieved through the use of a Tukey’s biweight loss function. 
An additional parameter controls the variability of feature weights across states, allowing the model to assign state-specific relevance to each feature. 
We illustrate in simulation how the method accurately recovers the true cluster sequence and reliably identifies relevant features, outperforming competing approaches, particularly in the presence of outliers. We conclude with two empirical applications, one on the number of conflict-related homicides in Kosovo in the period 1998–2000,  and another on macroeconomic performance of twelve European countries in the period 1949–2024.
}




\maketitle

\section{Introduction}

When performing dynamical clustering, it is often crucial to assess which variables contribute the most to the characterisation of the underlying structure. In fact, in many applications one is interested not only in segmenting a multivariate time series into homogeneous states, but also in understanding which variables drive its dynamics, possibly within each regime. An example is provided by jump models \citep[JM,][]{bemporad:2018}, which allow the estimation of persistent regimes without requiring a full parametric specification; 
as required by the related setting of hidden Markov models (HMM, \citet{bart:farc:penn:13,zucchini:2017}).
\citet{witten2010framework} introduce sparse $k$-means, a framework combining $k$-means with $\ell_1$ regularization, later extended by \citet{nystrup:2021} to the JM class. See also \citet{nystrup:2020}. 
Extensions of JM include models for mixed-type data with missing data imputation \citep{cortese:2025}, soft clustering variants \citep{aydinhan2024identifying}, and spatio-temporal extensions  \citep{cortese2026spatio}.

However, most existing approaches involve the same relevance for each feature in each cluster. This assumption is often restrictive, as different subsets of variables may govern different temporal regimes. Robustness is an additional key aspect, especially in the presence of noise and outliers, a common issue in real-world data.

In this work, we build on Clustering Objects on Subsets of Attributes (COSA) \citep{friedman2004clustering,kampert2017rcosa}, which is currently restricted to cross-sectional data, to provide a robust clustering approach for multivariate time series. COSA uses a dissimilarity-based formulation, where cluster-specific feature weights are entered directly into the dissimilarity measure. This results in a flexible, interpretable, and data-driven notion of variable relevance.
%

We thus introduce a robust Feature-Weighted Jump Model (FWJM) that jointly estimates a latent state sequence and a matrix of state-specific feature weights. 
%
 A hyperparameter regulates the frequency of transitions between latent states, while an entropy-based regularization parameter controls the distribution of state-specific feature weights.
%
%
%
The resulting model estimates state-conditional feature relevance through weighted combinations of feature-wise dissimilarities, with weights that are specific to each latent state and learned by minimising within-state dispersion. This allows different states to be characterized by different subsets of features, possibly overlapping. Robustness is achieved through the use of a Tukey-type loss function, which downweights extreme observations. 
%
%
%
Our model is distribution-free, and consequently flexible even in complex data scenarios. 

In order to assess the performance of the proposed method and to illustrate the advantage of its main features (robust loss function, cluster-dependent feature weights), we conduct an extensive simulation study, showing that the proposed method accurately recovers the latent state sequence and state-specific feature weights, outperforming competing methods, particularly in the presence of outliers.
We finally show two real data applications with slightly different characteristics in terms of latent persistence, sample size, and  dimensionality. 

The rest of the paper is organised as follows. Section \ref{sec:methodology} introduces the proposed methodology and proposes an efficient algorithm for estimation. Section \ref{sec:simstud} reports the simulation results, and Section \ref{sec:applications} illustrates the empirical applications. Section \ref{sec:conclusions} gives some concluding remarks.

\section{Methodology}
\label{sec:methodology}

We consider multivariate time series data collected in a $T\times P$ matrix denoted as $\boldsymbol{Y}$, where each row $\boldsymbol{y}_t=(y_{t1},\ldots,y_{tP})$ represents the observation vector at time $t$. We also indicate the time-specific cluster assignments with $\boldsymbol{s}=(s_1,\ldots,s_T)^\prime$, where $s_t\in\{1,\ldots,K\}$ for all $t$. 
Let also $\boldsymbol{W}$ denote a ${K\times P}$ matrix of feature weights, with generic element $w_{kp}$ being the weight of feature $p$ in cluster $k$; and $\sum_p w_{kp}=1$. 

We base the analyses on an appropriate dissimilarity matrix $D_{tt'}[\boldsymbol{W}]$ whose generic element corresponds to 
\begin{equation}
\label{eq:D}
\max\left\{\sum_{p=1}^P w_{s_t,p}d_{tt',p}, \sum_{p=1}^P w_{s_{t'},p}d_{tt',p}\right\}\,
\end{equation}
where the use of the maximum operator ensures that $D_{tt'}[\boldsymbol{W}]$ is well-defined for all pairs $(t,t')$, including those belonging to different clusters.

In \eqref{eq:D}, $d_{tt',p}$ denotes a (possibly robust) dissimilarity between $\boldsymbol{y}_t$ and $\boldsymbol{y}_{t'}$. 
To possibly accommodate the methodology for mixed-type measurements, we propose using a (modified) \citet{gower1971general} dissimilarity as follows. 
Partition the feature indices into
\begin{align*}
  \mathcal{C}&=\{p:\text{continuous}\},\\
  \mathcal{G}&=\{p:\text{categorical}\},\\
  \mathcal{O}&=\{p:\text{ordinal}\}.
\end{align*}
For any pair $(t,t^\prime)$ and each feature $p$, the classical  \cite{gower1971general} dissimilarity is
\begin{equation}
\label{eq:gower}
d_{tt^\prime,p} =
\begin{cases}
\displaystyle \frac{\lvert y_{tp}-y_{t^\prime p}\rvert}{\gamma_p}, 
  & p\in\mathcal{C},\\[1.2em]
\displaystyle \mathbb{I}(y_{t p}\neq y_{t^\prime p}), 
  & p\in\mathcal{G},\\[1.2em]
\displaystyle \frac{\bigl\lvert \mathrm{rank}_t(y_{t p}) - \mathrm{rank}_{t^\prime}(y_{t^\prime p})\bigr\rvert}{M_p - 1}, 
  & p\in\mathcal{O}.
\end{cases}
\end{equation}
Here, $\mathrm{rank}_t(y_{t p})$ is the rank among the $T$ values of the feature $p$, $M_p$ is the number of ordinal levels of the same feature, and $\gamma_p$ is 
the range for continuous features.

A robustified version is obtained with a slight modification of the dissimilarity for continuous features. We propose using Tukey's biweight loss as follows. 
\begin{enumerate}
    \item 
    For $p=1,\ldots,P$, 
    we first compute
    \begin{equation}
        \Delta_{tt',p}
        =
        \left|y_{tp} - y_{t^{\prime}p}\right|,
        \qquad t,t^{\prime}=1,\dots,T.
    \end{equation}

    \item 
    Let
    $
        \mathrm{MAD}_p
        =
        \mathrm{median}_{t}
        \left(
        \left|y_{tp} - \mathrm{median}_{t'}(y_{t'p})\right|
        \right).
   $
    Then the standardized differences are
    \begin{equation}
        u_{tt',p}
        =
        \frac{\Delta_{tt',p}}{\mathrm{MAD}_p}.
    \end{equation}

    \item 
    Compute
    \begin{equation}
    \label{eq:tukey_code}
        \rho\!\left(u_{tt',p}\right)
        =
        \begin{cases}
        \dfrac{c^2}{6}
        \left[
        1 -
        \left(
        1-\left(\dfrac{u_{tt',p}}{c}\right)^2
        \right)^3
        \right],
        & \text{if } \left|u_{tt',p}\right|\le c, \\[1em]
        \dfrac{c^2}{6},
        & \text{if } \left|u_{tt',p}\right|>c,
        \end{cases}
    \end{equation}
    with tuning constant $c = 4.685$.
    We select this value to guarantee approximately 95\% asymptotic efficiency at the normal model.

    \item Denote by
    \begin{equation}
        r_{tt',p}
        =
        \rho\!\left(u_{tt',p}\right),
        \qquad t,t'=1,\dots,T,
    \end{equation}
    the Tukey-transformed dissimilarities. 
    We finally apply a slice-wise normalization 
    \begin{equation}
    \label{eq:robust_gower}
         d_{tt',p}
        =
        \frac{r_{tt',p}}{\max_{t,t'} r_{tt',p} - \min_{t,t'} r_{tt',p}}.
    \end{equation}
\end{enumerate}
%

%


\subsection{Robust Feature Weighted Jump Models}

A FWJM is fitted by minimising the following objective function: 
%
\begin{equation}
    \label{eq:FWJM}
    f(\boldsymbol{s},\boldsymbol{W},\boldsymbol{m})=  
   \sum_t \sum_p w_{s_t,p}d_{tm_{s_t},p}+
   \zeta\sum_k\sum_p w_{kp}\log w_{kp}+\lambda\sum_{t=1}^{T-1} \mathbb{I}(s_t\neq s_{t+1}), 
\end{equation}
where $\boldsymbol{m}$ denotes a $K$ by $P$ matrix of medoids. 
That is, $m_k$ corresponds to one of the observations $y_{t}$ such that $s_t=k$. In \eqref{eq:FWJM}, with a slight abuse of notation, 
$d_{tm_{s_t},p}$ denotes the dissimilarity between $y_t$ and $m_{s_t}$, the medoid of the assigned cluster. Notably, for (at least) one of the cluster members, this dissimilarity will be zero. 

The first term is the main objective and leads to formal clustering: 
a small overall dissimilarity indicates that all observations are close to the medoid of their assigned cluster, and consequently to each other. 
Features with a high weight will contribute more to the determination of cluster assignments, as the first term is directly proportional to cluster-specific feature weights $w_{s_t,p}$. The term is a modification of the classical Partitioning Around Medoids (PAM) objective function, where we use possibly robustified dissimilarities, and class-specific feature weights. 

The second term is a penalty term used to promote less polarised weights. In fact, when $\zeta\to\infty$, the second term dominates and the minimum will be achieved when $w_{kp}=1/P$. 

The third term promotes persistence over time. Intuitively, it forces consecutive observations to stay in the same cluster with respect to observations that are far apart in time. 
%
When $\lambda = 0$, there is no temporal smoothing of the state sequence. 
As $\lambda$ increases, the penalty enforces persistence in the latent state sequence, discouraging transitions. For $\lambda \to \infty$, the model collapses to a single-state solution, i.e. a constant $s_t$. 
The range of $\lambda$ values for which the number of visited states is effectively equal to $K$ is approximately $[0,1]$ when using the metric defined in Equation \eqref{eq:gower}, as empirically confirmed in our simulation study of Section \ref{sec:simstud}. Nevertheless, in general, it is possible that for large penalty values some states are estimated as empty. Our estimation procedure, described in the next section, is fairly stable even when this occurs. Empty groups do not contribute to the objective function by definition, and the corresponding weights are estimated as $1/P$.

\subsection{Estimation and choice of penalty parameters}


We use a numerical optimisation procedure to minimise \eqref{eq:FWJM}. From an initial solution, 
we iterate three steps, updating each group of parameters separately and decreasing the overall objective function at each step. The procedure is stopped when, after a block of three steps, the improvement in the objective function is below a low tolerance value. 
%
%

   At the first step, holding the current weight matrix and cluster assignments fixed, medoids are directly updated in a greedy fashion: 
    \begin{equation}   \label{eq:updatemedoids_2}\boldsymbol{m}_k = \boldsymbol{y}_{t_k}, \,\,\text{s.t.},\,\,t_k =\underset{i \,:\, s_i = k}{\text{argmin}}\sum_{t \,:\, s_t = k}\sum_{p=1}^Pw_{kp}\, d_{t i,p},\end{equation}
for $k=1, \ldots, K$. 
In words, the observation within each cluster that minimises the weighted sums of dissimilarities to the observations in the same cluster is fixed as the current medoid. By construction, the first addend in \eqref{eq:FWJM} is therefore decreased or remains constant; while the other two addends remain constant as they do not depend on the medoid assignment.



At the second step, conditionally on $\boldsymbol{m}$ and the current weight matrix $\boldsymbol{W}$, we update the state sequence via dynamic programming. 
We do so by defining a value function recursively as
\begin{equation}
\begin{aligned}
\label{eq:Vfun}
V(T,k) &= D_{Tt_k}[\boldsymbol{W}], \\
V(t,k) &= D_{tt_k}[\boldsymbol{W}] 
+ \min_{j} \left\{ V(t+1,j) + \lambda \mathbb{I}(k \neq j) \right\}, 
\quad t = 1,\ldots,T-1,
\end{aligned}
\end{equation}
and then updating the optimal state sequence via backtracking
\begin{equation}
\begin{aligned}
\label{eq:s_fit}
s_1 &= \arg\min_{j} V(1,j), \\
s_t &= \arg\min_{j} \left\{ V(t,j) + \lambda \mathbb{I}(s_{t-1} \neq j) \right\}, 
\quad t = 2,\ldots,T.
\end{aligned}
\end{equation}
This approach is similar to the one proposed in \cite{nystrup:2020} for a slightly different objective function, and it can be shown with similar arguments that the sum of first and third term in \eqref{eq:FWJM} is not increased as a consequence of this update. The second addend remains constant as it is not a function of $\boldsymbol{s}$. 

Finally, at the third step, conditionally on 
$\boldsymbol{s}$ and $\boldsymbol{m}$, weights are updated in closed form as 
\begin{equation}
\label{eq:wkp_norm}
w_{kp}
=
\frac{
\exp\left(-\frac{S_{kp}}{\zeta}\right)
}{
\sum_{q=1}^P
\exp\left(-\frac{S_{kq}}{\zeta}\right)
}
\end{equation}
with
$$
S_{kp}
=
\sum_{t:s_t=k} d_{t m_k,p}.
$$
We show in Section S2 of the Supplementary material how this closed form solution minimises the objective function as a function of the weight matrix. 

The procedure is iterated until convergence of Eq. \eqref{eq:FWJM} or until a maximum number of iterations is reached. 
Since at each step of the algorithm the objective function is decreased or remains constant, at convergence we will have found a local minimum. 
%
%
%
Algorithm~\ref{alg:FWJM} provides a synthetic description of the estimation procedure.
\begin{algorithm}[t]
\caption{Robust Feature Weighted Jump Model (FWJM) fitting}
\label{alg:FWJM}
\textbf{Input:} Data $\boldsymbol{y}_1,\ldots,\boldsymbol{y}_T \in \mathbb{R}^P$, number of states $K$, hyperparameters $\lambda$ and $\zeta$

\begin{algorithmic}[1]

\State \textbf{Initialisation:}
\begin{enumerate}
    \item[(a)] Initialise state sequence $\boldsymbol{s}$
    \item[(b)] 
    Initialise weight matrix $\boldsymbol{W}$
\end{enumerate}

\State \textbf{Loop}. Repeat until convergence:
\begin{enumerate}

    \item [(i)] Compute robust differences $d_{tt^\prime,p}$ as in \eqref{eq:gower} and \eqref{eq:robust_gower}
        \item[(ii)] Compute pairwise dissimilarities $D_{tt'}[\boldsymbol{W}]$ using
        \eqref{eq:D}

        \item[(iii)] Find medoids $\boldsymbol{m}$ solving \eqref{eq:updatemedoids_2}  
            \item[(iv)] Update the state sequence $\boldsymbol{s}=(s_1,\ldots,s_T)^\prime$ through \eqref{eq:Vfun} and \eqref{eq:s_fit}

        \item[(v)] Update feature weights $\boldsymbol{W}$ using \eqref{eq:wkp_norm}
    \end{enumerate}

\State \textbf{Return}: medoids $\boldsymbol{m}$, state sequence $\boldsymbol{s}$, feature weights matrix $\boldsymbol{W}$

\end{algorithmic}
\end{algorithm}

We highlight that the proposed estimation algorithm relies on an alternating
optimization scheme, and due to the highly non-convex nature of the objective function, the algorithm is not guaranteed to reach the global optimum in general. The classical solution in computational statistics involves a multi-start strategy. We use here the same procedure suggested in \cite{nystrup:2020}. In our implementation ten different initial solutions are compared. 
Regarding the choice of the hyperparameters controlling the state persistence ($\lambda$) and the distribution of feature weights ($\zeta$) one may rely on cross-validation procedures when ground truth labels are available at least for a subset of the data, or on task-specific objective criteria (e.g., as in \cite{nystrup2019multi}). In general, we propose using internal clustering validation measures, such as the Silhouette index \citep{rousseeuw1987silhouettes}. The procedure involves computing the Silhouettes corresponding to different penalty parameter specifications, and selecting the choices yielding the best cluster configuration. This is precisely what we will do in the applications presented in Section \ref{sec:applications}. 

Reproducible code for estimation and hyperparameter selection is available in the GitHub repository:
\url{https://github.com/FedericoCortese/FWJM}.

\section{Simulation Study}
\label{sec:simstud}

In order to assess the accuracy of the proposed method, we consider four simulation scenarios. Scenario A represents a classical setting with $T=1\,000$ observations and $P=5$ features. Scenario B corresponds to a \textit{large-$P$, small-$T$} setting with $T=P=50$. Scenario C considers a high-dimensional setting with $T=1\,000$ and $P=50$, while scenario D corresponds to a very challenging \textit{small-$P$, small-$T$} setting with $T=50$ and $P=5$.

We generate data $\boldsymbol{Y}$ from a multivariate Student-$t$ HMM with $\nu=3$ degrees of freedom and $K=2,3,$ or $4$ latent states. The state-dependent distributions are characterized by centroids $\mu_k$,  and covariance matrices with unit variances and constant pairwise correlation $\rho=0.2$ across all features and states. 

We set the self-transition probabilities $\pi_{ii}=\mathbb{P}(s_t=i \mid s_{t-1}=i)$, $i=1,\ldots,K$, to $0.99$ in scenarios A and C, while for scenarios B and D they are set to $0.95$ when $K=2$, to $0.90$ when $K=3$, and to $0.80$ when $K=4$, in order to ensure an adequate number of visits to all latent states. Off-diagonal transition probabilities are defined as
$
\pi_{ij}=\frac{1-\pi_{ii}}{K-1},
\, i \neq j.
$

When $K=2$, the centroid are $(-0.5,0.5)$ in scenarios A and C, and $(1,1)$ in the other two scenarios, while
for $K=3$, they are $(-1,0,1)$ in scenarios A and C and $(-2,0,2)$ in scenarios B and D. 
When $K=4$, the centroids are $(-1.5,-0.5,0.5,1.5)$ in scenarios A and C and $(-3,-1,1,3)$ in scenarios B and D. 
We choose these configurations to control cluster separation across scenarios: adjacent centroids differ by 1 in scenarios A and C, and by 2 in scenarios B and D, the latter therefore exhibiting greater separation to compensate for the increased difficulty induced by the small-sample.

For ease of reference, Table~\ref{tab:simstud_params} summarizes the data-generating parameters that vary across scenarios, including sample size $T$, number of features $P$, self-transition probabilities $\pi_{ii}$, and state-specific centroids $\mu_p$.
\begin{table}[htbp]
\centering
\caption{Summary of varying data-generating parameters across scenarios: self-transition probabilities $\pi_{ii}$, $i=1,\ldots,K$, of the latent Markov process and centroids $\mu_p$, $p=1,\ldots,P$, for different values of $K$.}
\label{tab:simstud_params}

\begin{tabular}{llcccc}
\hline
 & $K$ & Scenario A & Scenario B & Scenario C & Scenario D \\
\hline

\multirow{3}{*}{$T$}
& $2$ &   &   &   &   \\
& $3$ & $1\,000$ & 50 & $1\,000$ & 50 \\
& $4$ &   &   &   &   \\
\hline

\multirow{3}{*}{$P$}
& $2$ &   &   &   &   \\
& $3$ & 5 & 50 & 50 & 5 \\
& $4$ &   &   &   &   \\
\hline

\multirow{3}{*}{$\pi_{ii}$}
& $2$ &  & 0.95 &  & 0.95 \\
& $3$ & 0.99 & 0.90 & 0.99 & 0.90 \\
& $4$ &  & 0.80 &  & 0.80 \\
\hline

\multirow{3}{*}{$\mu_p$}
& $2$ 
& $-0.5,\,0.5$ 
& $-1,\,1$ 
& $-0.5,\,0.5$ 
& $-1,\,1$ \\

& $3$ 
& $-1,\,0,\,1$ 
& $-2,\,0,\,2$ 
& $-1,\,0,\,1$ 
& $-2,\,0,\,2$ \\

& $4$ 
& $-1.5,\,-0.5,\,0.5,\,1.5$ 
& $-3,\,-1,\,1,\,3$ 
& $-1.5,\,-0.5,\,0.5,\,1.5$ 
& $-3,\,-1,\,1,\,3$ \\

\hline

\end{tabular}
\end{table}

Similarly to the simulation studies in \citet{malsiner2016model} and \citet{nystrup:2021}, we artificially introduce non-informative features by replacing a subset of variables with random noise. Specifically, these are generated as independent draws from a uniform distribution $\mathcal{U}(y_m, y_M)$, where $y_m = \min \boldsymbol{Y}$ and $y_M = \max \boldsymbol{Y}$. 
Hence, whenever a feature is not relevant for a given state, or is purely noise, its values are sampled from this uniform distribution.
In scenarios A and D, each state is associated with two (when $K=2$) and one ($K=3,4$) state-specific feature and one ($K=2,4)$ or two ($K=3$) globally informative feature shared across all states. In scenarios B and C, each state is associated with a block of ten ($K=2$) or five ($K=3,4$) state-specific informative features, five ($K=2,4$) or ten ($K=3$) shared features, and the remaining features are purely noisy.

To evaluate robustness, we consider two settings, one with contamination level  $\alpha=5\%$, the other with no contamination, $\alpha=0\%$. We generate contaminated observations from $\mathcal{U}(y_m - q, y_M + q)$, with $q = 30$: given the heavy-tailed nature of the Student-$t$ distribution with $\nu=3$, we empirically observe that this choice helps distinguish true outliers from observations arising from the tails of the distribution.

We estimate FWJM and its non-robust counterpart (nr-FWJM), obtained without considering the Tukey correction, over a grid of values for $\lambda$ and $\zeta$. 
When $P=5$, we consider $\zeta \in \{0.5, 1, 5, 10, 25, 50, 100\}$, whereas for $P=50$ we consider $\zeta \in \{0.05, 0.1, 0.2, 0.5, 1, 5, 10\}$, since smaller values are expected to be optimal in the presence of many noisy features. 
Similarly, we consider $\lambda \in \{0, 0.25, 0.5, 0.75, 1\}$ when $T=1\,000$, and $\lambda \in \{0, 0.05, 0.1, 0.2, 0.5\}$ when $T=50$, as smaller values are expected to be optimal for shorter time series. 
Each configuration is replicated 200 times.

The FWJM is compared against several benchmarks, including nr-FWJM, $k$-means, the JM of \cite{nystrup:2020}, the SJM of \cite{nystrup:2021}, a multivariate Student-$t$ HMM, and the COSA method of \cite{friedman2004clustering}.

We evaluate performance in terms of clustering and estimation accuracy. Specifically, we report the median balanced accuracy (BAC) and adjusted Rand index (ARI) between the true and estimated state sequences, and the root mean squared error (RMSE) between true centroids and estimated prototypes. For the FWJM, we additionally show the medians of the estimated feature weights across scenarios.

\subsection{Results}

Table \ref{tab:results_noout} summarizes the results for the simulation settings without contamination. Overall, the proposal consistently provides the best clustering performance across all scenarios and values of $K$, achieving ARI and BAC values close or equal to 1 in most settings, while maintaining competitive RMSE values. Its advantage is particularly pronounced in scenarios A and C ($T=1\,000)$,  where competing approaches often fail to recover the latent state structure. 
Among the competing methods, COSA generally provides the strongest alternative performance, especially in scenario B and, to a lesser extent, in scenario C. The Stud-$t$ HMM also achieves good performance in some settings, particularly for smaller values of $K$. 
In contrast, SJM, JM, and $k$-means exhibit consistently poor performance across most scenarios.%
\begin{sidewaystable}[htbp]
\centering
\caption{
Performance comparison across four scenarios with $K=2, K=3$ and $K=4$ latent states and no contamination. 
Methods include the feature-weighted jump model (FWJM), its non-robust version (nr-FWJM), the sparse jump model (SJM), the jump model (JM), $k$-means, Clustering Objects on Subsets of Attributes (COSA), and a Student-$t$ hidden Markov model (Stud-$t$ HMM). Results are reported in terms of median adjusted Rand index (ARI) and balanced accuracy (BAC), with standard deviations in parentheses, and root mean squared errors (RMSE) between true centroids and estimated medoids. Bold values indicate the best performance, corresponding to the highest ARI and BAC and the lowest RMSE.
}
\renewcommand{\arraystretch}{1.3}
\begin{tabular}{llcccccccccccc}
\hline
 &  & \multicolumn{3}{c}{Scenario A} & \multicolumn{3}{c}{Scenario B} & \multicolumn{3}{c}{Scenario C} & \multicolumn{3}{c}{Scenario D} \\
\toprule
$K$ & Method & ARI & BAC & RMSE & ARI & BAC & RMSE & ARI & BAC & RMSE & ARI & BAC & RMSE \\
\toprule

\multirow{7}{*}{2}
& FWJM
& \celltwo{\textbf{1.00}}{0.10}
& \celltwo{\textbf{1.00}}{0.05}
& 0.29
& \celltwo{\textbf{1.00}}{0.39}
& \celltwo{\textbf{1.00}}{0.16}
& 0.80
& \celltwo{\textbf{1.00}}{0.09}
& \celltwo{\textbf{1.00}}{0.02}
& 0.15
& \celltwo{\textbf{0.92}}{0.42}
& \celltwo{0.99}{0.18}
& 0.77 \\

& nr-FWJM
& \celltwo{\textbf{1.00}}{0.12}
& \celltwo{\textbf{1.00}}{0.04}
& 0.35
& \celltwo{\textbf{1.00}}{0.36}
& \celltwo{\textbf{1.00}}{0.12}
& 0.75
& \celltwo{\textbf{1.00}}{0.00}
& \celltwo{\textbf{1.00}}{0.00}
& \textbf{0.11}
& \celltwo{\textbf{0.92}}{0.35}
& \celltwo{\textbf{1.00}}{0.11}
& 0.69 \\

& SJM
& \celltwo{0.09}{0.07}
& \celltwo{0.72}{0.02}
& 0.83
& \celltwo{0.01}{0.22}
& \celltwo{0.64}{0.17}
& \textbf{0.62}
& \celltwo{0.00}{0.09}
& \celltwo{0.52}{0.09}
& 0.71
& \celltwo{0.05}{0.21}
& \celltwo{0.75}{0.15}
& 0.87 \\

& JM
& \celltwo{0.08}{0.08}
& \celltwo{0.72}{0.06}
& 0.89
& \celltwo{0.01}{0.31}
& \celltwo{0.64}{0.18}
& \textbf{0.62}
& \celltwo{0.00}{0.21}
& \celltwo{0.52}{0.13}
& 0.63
& \celltwo{0.05}{0.21}
& \celltwo{0.75}{0.14}
& 0.86 \\

& $k$-means
& \celltwo{0.07}{0.08}
& \celltwo{0.72}{0.07}
& 0.90
& \celltwo{0.01}{0.25}
& \celltwo{0.62}{0.15}
& 0.64
& \celltwo{0.00}{0.21}
& \celltwo{0.51}{0.13}
& 0.62
& \celltwo{0.05}{0.17}
& \celltwo{0.74}{0.14}
& 0.85 \\

& COSA
& \celltwo{0.82}{0.13}
& \celltwo{0.96}{0.03}
& \textbf{0.17}
& \celltwo{0.92}{0.34}
& \celltwo{0.99}{0.09}
& 0.69
& \celltwo{0.98}{0.07}
& \celltwo{\textbf{1.00}}{0.05}
& 0.16
& \celltwo{0.63}{0.31}
& \celltwo{0.92}{0.11}
& \textbf{0.68} \\

& Stud-$t$ HMM
& \celltwo{0.22}{0.22}
& \celltwo{0.70}{0.08}
& 4.09
& \celltwo{0.70}{0.40}
& \celltwo{0.92}{0.19}
& 2.01
& \celltwo{0.98}{0.26}
& \celltwo{0.99}{0.13}
& 3.90
& \celltwo{0.47}{0.40}
& \celltwo{0.83}{0.19}
& 1.61 \\

\toprule

\multirow{7}{*}{3}
& FWJM
& \celltwo{\textbf{1.00}}{0.10}
& \celltwo{\textbf{1.00}}{0.06}
& 2.08
& \celltwo{\textbf{1.00}}{0.21}
& \celltwo{\textbf{1.00}}{0.08}
& 0.74
& \celltwo{\textbf{1.00}}{0.14}
& \celltwo{\textbf{1.00}}{0.05}
& 0.67
& \celltwo{0.93}{0.25}
& \celltwo{0.98}{0.09}
& 1.13 \\

& nr-FWJM
& \celltwo{\textbf{1.00}}{0.05}
& \celltwo{\textbf{1.00}}{0.06}
& 1.34
& \celltwo{\textbf{1.00}}{0.11}
& \celltwo{\textbf{1.00}}{0.07}
& 0.77
& \celltwo{\textbf{1.00}}{0.10}
& \celltwo{\textbf{1.00}}{0.07}
& 1.49
& \celltwo{\textbf{0.94}}{0.10}
& \celltwo{\textbf{0.99}}{0.07}
& 1.24 \\

& SJM
& \celltwo{0.07}{0.05}
& \celltwo{0.65}{0.02}
& 1.36
& \celltwo{0.11}{0.25}
& \celltwo{0.68}{0.11}
& 0.82
& \celltwo{0.00}{0.08}
& \celltwo{0.52}{0.04}
& 0.95
& \celltwo{0.18}{0.18}
& \celltwo{0.68}{0.08}
& 1.13 \\

& JM
& \celltwo{0.07}{0.05}
& \celltwo{0.59}{0.02}
& 1.47
& \celltwo{0.11}{0.25}
& \celltwo{0.67}{0.10}
& 0.82
& \celltwo{0.00}{0.10}
& \celltwo{0.51}{0.05}
& 0.92
& \celltwo{0.18}{0.20}
& \celltwo{0.68}{0.08}
& 1.14 \\

& $k$-means
& \celltwo{0.07}{0.05}
& \celltwo{0.59}{0.02}
& 1.47
& \celltwo{0.08}{0.21}
& \celltwo{0.67}{0.10}
& 0.85
& \celltwo{0.00}{0.10}
& \celltwo{0.51}{0.05}
& 0.93
& \celltwo{0.15}{0.18}
& \celltwo{0.67}{0.08}
& 1.14 \\

& COSA
& \celltwo{0.58}{0.12}
& \celltwo{0.88}{0.05}
& \textbf{0.51}
& \celltwo{0.91}{0.22}
& \celltwo{0.98}{0.07}
& \textbf{0.68}
& \celltwo{0.90}{0.10}
& \celltwo{0.97}{0.05}
& \textbf{0.63}
& \celltwo{0.70}{0.21}
& \celltwo{0.93}{0.09}
& \textbf{0.99} \\

& Stud-$t$ HMM
& \celltwo{0.57}{0.17}
& \celltwo{0.77}{0.05}
& 4.57
& \celltwo{0.62}{0.24}
& \celltwo{0.82}{0.13}
& 2.66
& \celltwo{0.74}{0.19}
& \celltwo{0.79}{0.11}
& 4.20
& \celltwo{0.66}{0.26}
& \celltwo{0.79}{0.12}
& 2.10 \\
\toprule

\multirow{7}{*}{4}
& FWJM
& \celltwo{0.99}{0.10}
& \celltwo{\textbf{1.00}}{0.06}
& \textbf{1.04}
& \celltwo{0.90}{0.18}
& \celltwo{0.93}{0.08}
& 0.86
& \celltwo{\textbf{1.00}}{0.15}
& \celltwo{\textbf{1.00}}{0.05}
& 1.15
& \celltwo{0.78}{0.24}
& \celltwo{0.85}{0.10}
& 1.61 \\

& nr-FWJM
& \celltwo{\textbf{1.00}}{0.04}
& \celltwo{\textbf{1.00}}{0.06}
& 1.44
& \celltwo{\textbf{0.94}}{0.11}
& \celltwo{0.93}{0.08}
& 0.75
& \celltwo{\textbf{1.00}}{0.10}
& \celltwo{\textbf{1.00}}{0.08}
& 1.71
& \celltwo{\textbf{0.85}}{0.16}
& \celltwo{\textbf{0.89}}{0.09}
& 1.78 \\

& SJM
& \celltwo{0.05}{0.04}
& \celltwo{0.62}{0.02}
& 1.76
& \celltwo{0.12}{0.15}
& \celltwo{0.66}{0.07}
& 0.87
& \celltwo{0.00}{0.11}
& \celltwo{0.52}{0.03}
& \textbf{1.14}
& \celltwo{0.10}{0.12}
& \celltwo{0.66}{0.06}
& 1.18 \\

& JM
& \celltwo{0.05}{0.04}
& \celltwo{0.57}{0.02}
& 2.20
& \celltwo{0.11}{0.17}
& \celltwo{0.66}{0.07}
& 0.87
& \celltwo{0.00}{0.06}
& \celltwo{0.52}{0.03}
& 1.17
& \celltwo{0.10}{0.12}
& \celltwo{0.66}{0.06}
& 1.18 \\

& $k$-means
& \celltwo{0.05}{0.04}
& \celltwo{0.57}{0.02}
& 2.20
& \celltwo{0.10}{0.16}
& \celltwo{0.65}{0.07}
& 0.87
& \celltwo{-0.00}{0.06}
& \celltwo{0.52}{0.03}
& 1.17
& \celltwo{0.09}{0.10}
& \celltwo{0.64}{0.06}
& 1.20 \\

& COSA
& \celltwo{0.46}{0.13}
& \celltwo{0.80}{0.06}
& 1.31
& \celltwo{0.87}{0.16}
& \celltwo{\textbf{0.96}}{0.06}
& \textbf{0.63}
& \celltwo{0.89}{0.13}
& \celltwo{0.97}{0.06}
& 1.19
& \celltwo{0.51}{0.18}
& \celltwo{0.82}{0.09}
& \textbf{1.12} \\

& Stud-$t$ HMM
& \celltwo{0.38}{0.11}
& \celltwo{0.66}{0.05}
& 5.54
& \celltwo{0.34}{0.17}
& \celltwo{0.72}{0.07}
& 3.20
& \celltwo{0.54}{0.23}
& \celltwo{0.72}{0.12}
& 4.78
& \celltwo{0.37}{0.18}
& \celltwo{0.71}{0.08}
& 3.28 \\

\bottomrule

\end{tabular}
\label{tab:results_noout}
\end{sidewaystable}

Table \ref{tab:results_main} reports the results under 5\% contamination. Overall, FWJM consistently achieves the best clustering performance across nearly all scenarios and values of $K$. The advantage of the robust formulation over its non-robust counterpart becomes particularly evident under contamination, with large performance gaps observed across several settings (e.g., ARI equal to 0.99 versus 0.63 in scenario A with $K=3$, and 0.99 versus 0.67 for $K=4$). 
COSA generally provides the strongest alternative performance in scenario B, where the shorter time dimension reduces the benefit of temporal persistence and allows COSA to achieve clustering accuracies comparable to FWJM in some cases, sometimes at the expense of larger RMSE values. The Stud-$t$ HMM attains moderate performance in some settings, especially in scenario C, but remains systematically inferior to FWJM.
As before, SJM, JM, and $k$-means perform poorly across all settings, with ARI values frequently close to zero and BAC values near random-assignment levels. 
\begin{sidewaystable}[htbp]
\centering
\caption{
Performance comparison across four scenarios with $K=2, K=3$ and $K=4$ latent states and 5\% contamination. 
Methods include the feature-weighted jump model (FWJM), its non-robust version (nr-FWJM), the sparse jump model (SJM), the jump model (JM), $k$-means, Clustering Objects on Subsets of Attributes (COSA), and a Student-$t$ hidden Markov model (Stud-$t$ HMM). Results are reported in terms of median adjusted Rand index (ARI) and balanced accuracy (BAC), with standard deviations in parentheses, and root mean squared errors (RMSE) between true centroids and estimated medoids. Bold values indicate the best performance, corresponding to the highest ARI and BAC and the lowest RMSE.
}
\renewcommand{\arraystretch}{1.3}
\begin{tabular}{llcccccccccccc}
\hline
 &  & \multicolumn{3}{c}{Scenario A} & \multicolumn{3}{c}{Scenario B} & \multicolumn{3}{c}{Scenario C} & \multicolumn{3}{c}{Scenario D} \\
\toprule
$K$ & Method & ARI & BAC & RMSE & ARI & BAC & RMSE & ARI & BAC & RMSE & ARI & BAC & RMSE \\
\toprule

\multirow{7}{*}{2}
& FWJM
& \celltwo{\textbf{0.99}}{0.13}
& \celltwo{\textbf{1.00}}{0.08}
& \textbf{0.32}
& \celltwo{\textbf{0.37}}{0.44}
& \celltwo{\textbf{0.83}}{0.24}
& \textbf{5.94}
& \celltwo{\textbf{1.00}}{0.10}
& \celltwo{\textbf{1.00}}{0.05}
& \textbf{0.23}
& \celltwo{0.01}{0.24}
& \celltwo{\textbf{0.54}}{0.17}
& 9.46 \\

& nr-FWJM
& \celltwo{0.96}{0.41}
& \celltwo{0.99}{0.21}
& 2.91
& \celltwo{-0.01}{0.24}
& \celltwo{0.49}{0.15}
& 8.03
& \celltwo{0.99}{0.14}
& \celltwo{1.00}{0.08}
& 0.37
& \celltwo{\textbf{0.08}}{0.45}
& \celltwo{0.50}{0.22}
& \textbf{8.29} \\

& SJM
& \celltwo{0.05}{0.07}
& \celltwo{0.68}{0.09}
& 3.52
& \celltwo{-0.01}{0.05}
& \celltwo{0.49}{0.05}
& 6.62
& \celltwo{0.00}{0.08}
& \celltwo{0.52}{0.08}
& 0.83
& \celltwo{-0.00}{0.06}
& \celltwo{0.49}{0.03}
& 10.50 \\

& JM
& \celltwo{0.02}{0.07}
& \celltwo{0.68}{0.09}
& 4.19
& \celltwo{-0.01}{0.04}
& \celltwo{0.49}{0.03}
& 8.12
& \celltwo{-0.00}{0.17}
& \celltwo{0.52}{0.11}
& 0.80
& \celltwo{-0.01}{0.06}
& \celltwo{0.49}{0.03}
& 11.60 \\

& $k$-means
& \celltwo{0.02}{0.07}
& \celltwo{0.68}{0.09}
& 4.16
& \celltwo{-0.02}{0.04}
& \celltwo{0.49}{0.02}
& 8.01
& \celltwo{-0.00}{0.17}
& \celltwo{0.51}{0.11}
& 0.77
& \celltwo{-0.01}{0.05}
& \celltwo{0.49}{0.03}
& 11.60 \\

& COSA
& \celltwo{0.08}{0.16}
& \celltwo{0.70}{0.10}
& 3.51
& \celltwo{-0.00}{0.25}
& \celltwo{0.50}{0.15}
& 8.11
& \celltwo{0.87}{0.44}
& \celltwo{0.97}{0.23}
& 7.53
& \celltwo{0.00}{0.23}
& \celltwo{0.51}{0.15}
& 10.8 \\

& Stud-$t$ HMM
& \celltwo{0.20}{0.18}
& \celltwo{0.69}{0.07}
& 4.17
& \celltwo{-0.02}{0.06}
& \celltwo{0.49}{0.04}
& 15.80
& \celltwo{0.96}{0.37}
& \celltwo{0.99}{0.18}
& 4.43
& \celltwo{0.00}{0.34}
& \celltwo{0.51}{0.19}
& 13.50 \\

\toprule

\multirow{7}{*}{3}
& FWJM
& \celltwo{\textbf{0.99}}{0.10}
& \celltwo{\textbf{1.00}}{0.06}
& 2.87
& \celltwo{\textbf{0.66}}{0.34}
& \celltwo{\textbf{0.77}}{0.12}
& \textbf{7.74}
& \celltwo{\textbf{1.00}}{0.13}
& \celltwo{\textbf{1.00}}{0.09}
& 2.59
& \celltwo{\textbf{0.68}}{0.24}
& \celltwo{\textbf{0.77}}{0.06}
& 10.4 \\

& nr-FWJM
& \celltwo{0.63}{0.22}
& \celltwo{0.78}{0.09}
& 11.9
& \celltwo{0.44}{0.33}
& \celltwo{0.73}{0.13}
& 8.36
& \celltwo{0.59}{0.21}
& \celltwo{0.77}{0.11}
& 6.77
& \celltwo{0.48}{0.31}
& \celltwo{0.74}{0.12}
& 11.7 \\

& SJM
& \celltwo{0.03}{0.05}
& \celltwo{0.60}{0.06}
& 4.97
& \celltwo{-0.00}{0.09}
& \celltwo{0.52}{0.05}
& 6.99
& \celltwo{0.00}{0.06}
& \celltwo{0.51}{0.03}
& 1.30
& \celltwo{0.01}{0.19}
& \celltwo{0.52}{0.08}
& 11.10 \\

& JM
& \celltwo{0.03}{0.05}
& \celltwo{0.59}{0.05}
& 5.10
& \celltwo{-0.01}{0.04}
& \celltwo{0.51}{0.03}
& 9.31
& \celltwo{0.00}{0.07}
& \celltwo{0.51}{0.04}
& \textbf{1.22}
& \celltwo{-0.00}{0.06}
& \celltwo{0.51}{0.03}
& 13.40 \\

& $k$-means
& \celltwo{0.02}{0.05}
& \celltwo{0.59}{0.05}
& 5.10
& \celltwo{-0.01}{0.04}
& \celltwo{0.51}{0.02}
& 9.36
& \celltwo{0.00}{0.07}
& \celltwo{0.51}{0.04}
& \textbf{1.22}
& \celltwo{-0.00}{0.06}
& \celltwo{0.51}{0.03}
& 13.40 \\

& COSA
& \celltwo{0.42}{0.15}
& \celltwo{0.72}{0.05}
& 11.9
& \celltwo{0.63}{0.18}
& \celltwo{0.76}{0.05}
& 6.98
& \celltwo{0.58}{0.11}
& \celltwo{0.75}{0.02}
& 9.06
& \celltwo{0.55}{0.23}
& \celltwo{0.75}{0.08}
& \textbf{9.96} \\

& Stud-$t$ HMM
& \celltwo{0.51}{0.19}
& \celltwo{0.76}{0.06}
& 4.59
& \celltwo{-0.00}{0.23}
& \celltwo{0.51}{0.08}
& 18.00
& \celltwo{0.66}{0.20}
& \celltwo{0.78}{0.09}
& 4.91
& \celltwo{0.58}{0.32}
& \celltwo{0.75}{0.12}
& 13.80 \\

\toprule

\multirow{7}{*}{4}
& FWJM
& \celltwo{\textbf{0.99}}{0.14}
& \celltwo{\textbf{1.00}}{0.08}
& 3.83
& \celltwo{0.50}{0.19}
& \celltwo{0.68}{0.06}
& 7.25
& \celltwo{\textbf{0.99}}{0.15}
& \celltwo{\textbf{1.00}}{0.07}
& 2.80
& \celltwo{\textbf{0.44}}{0.24}
& \celltwo{\textbf{0.70}}{0.08}
& 9.54 \\

& nr-FWJM
& \celltwo{0.67}{0.17}
& \celltwo{0.80}{0.08}
& 11.1
& \celltwo{0.43}{0.18}
& \celltwo{0.68}{0.06}
& 7.47
& \celltwo{0.63}{0.18}
& \celltwo{0.78}{0.08}
& 8.17
& \celltwo{0.36}{0.19}
& \celltwo{0.66}{0.06}
& 10.3 \\

& SJM
& \celltwo{0.02}{0.04}
& \celltwo{0.56}{0.04}
& 5.56
& \celltwo{0.01}{0.11}
& \celltwo{0.56}{0.05}
& 5.80
& \celltwo{0.00}{0.05}
& \celltwo{0.52}{0.03}
& 1.56
& \celltwo{0.07}{0.16}
& \celltwo{0.60}{0.06}
& 8.55 \\

& JM
& \celltwo{0.01}{0.04}
& \celltwo{0.54}{0.03}
& 5.70
& \celltwo{-0.00}{0.04}
& \celltwo{0.51}{0.02}
& 8.85
& \celltwo{0.00}{0.05}
& \celltwo{0.52}{0.03}
& \textbf{1.47}
& \celltwo{0.01}{0.09}
& \celltwo{0.53}{0.05}
& 11.40 \\

& $k$-means
& \celltwo{0.01}{0.04}
& \celltwo{0.54}{0.03}
& 5.45
& \celltwo{-0.00}{0.03}
& \celltwo{0.51}{0.02}
& 8.86
& \celltwo{0.00}{0.04}
& \celltwo{0.52}{0.03}
& 1.53
& \celltwo{0.01}{0.09}
& \celltwo{0.53}{0.05}
& 11.40 \\

& COSA
& \celltwo{0.41}{0.18}
& \celltwo{0.69}{0.07}
& 9.86
& \celltwo{\textbf{0.66}}{0.15}
& \celltwo{\textbf{0.80}}{0.07}
& \textbf{5.96}
& \celltwo{0.74}{0.13}
& \celltwo{0.82}{0.04}
& 7.01
& \celltwo{0.40}{0.20}
& \celltwo{\textbf{0.70}}{0.08}
& \textbf{8.44} \\

& Stud-$t$ HMM
& \celltwo{0.37}{0.10}
& \celltwo{0.64}{0.04}
& 5.86
& \celltwo{0.08}{0.18}
& \celltwo{0.58}{0.07}
& 18.00
& \celltwo{0.42}{0.20}
& \celltwo{0.68}{0.09}
& 8.85
& \celltwo{0.31}{0.22}
& \celltwo{0.66}{0.07}
& 14.80 \\

\bottomrule


\end{tabular}
\label{tab:results_main}
\end{sidewaystable}

We now turn to the optimal values of the hyperparameters $\zeta$ and $\lambda$.
%
For the persistence parameter $\lambda$, larger values are generally preferred when the time horizon $T$ is large, as in scenarios A and C, where temporal information is more informative and persistent state sequences can be reliably estimated. In these settings, the optimal values are typically around $\lambda=0.25$--$0.75$, with values close to $0.5$ most frequently selected. By contrast, when $T$ is smaller, lower persistence penalties are systematically favored, with optimal values usually between $\lambda=0.05$ and $\lambda=0.20$. 
A similar dependence is observed for the hyperparameter $\zeta$. When most features are informative and the temporal signal is strong, larger values of $\zeta$ are preferred, typically ranging between $10$ and $100$, as observed in scenario A. Conversely, in settings characterized by shorter series and a larger proportion of noisy features, smaller values of $\zeta$ are systematically selected. In particular, in scenarios B and D the optimal values are often between $0.2$ and $5$, with the smallest values generally preferred for larger values of $K$ and in contaminated settings. 

\begin{figure}[!htbp]  
    \centering
    \includegraphics[width=.48\linewidth]{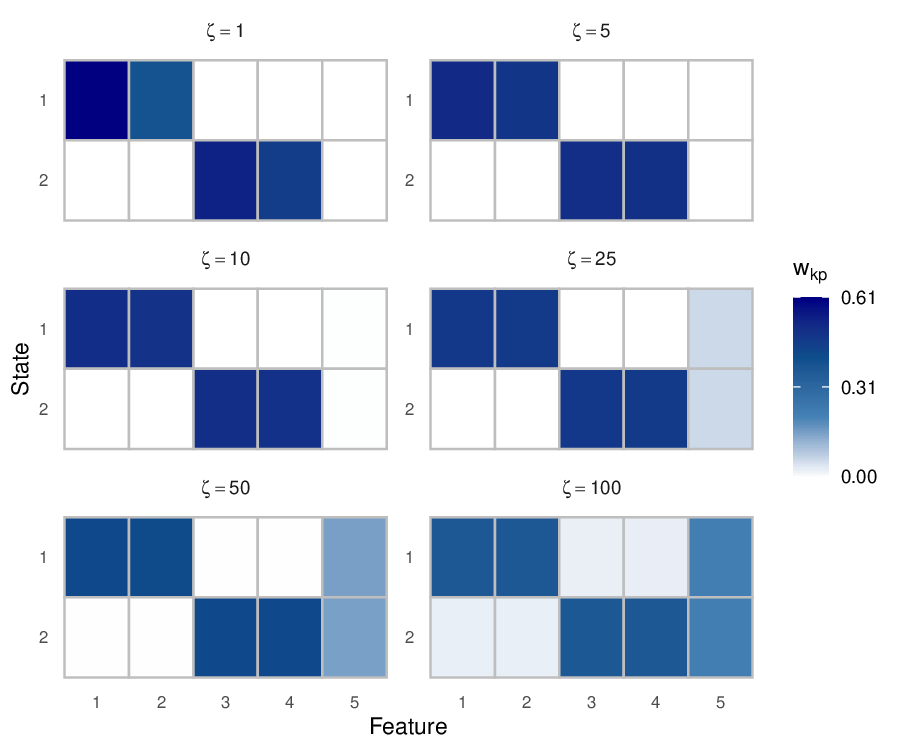}
    \includegraphics[width=.48\linewidth]{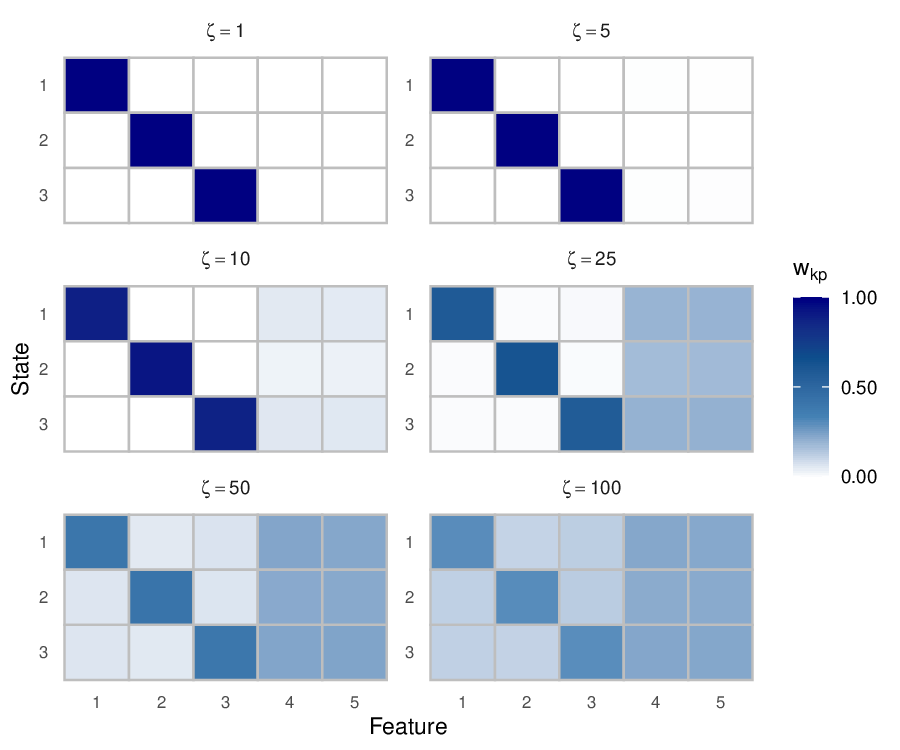}
    \includegraphics[width=.48\linewidth]{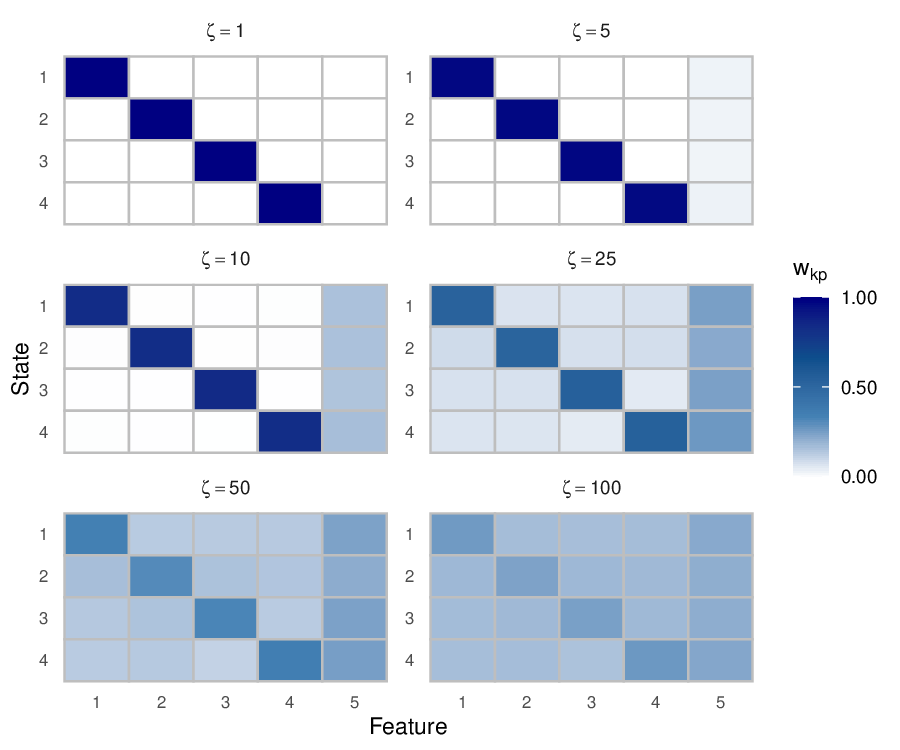}
    \caption{Normalized median weight matrices for FWJM under scenario A with $5\%$ contamination, for $K=2$ (top-left) $K=3$ and (top-right) $K=4$ (bottom) latent states.}
    \label{fig:est_W_scenA_out}
\end{figure}

The main contribution of FWJM lies in estimating state-dependent feature weights. 
%
%
Therefore, in Figure \ref{fig:est_W_scenA_out} we report the median estimated feature weights in scenario A for $K=2, 3$ and $4$, with 5\% contamination, and with $\lambda$ fixed at its optimal value according to ARI and BAC. The corresponding results in the absence of contamination are qualitatively identical and are therefore omitted for brevity.
Results show that the method assigns high weights to features that are informative for discriminating specific states.
Values of $\zeta$ between 25 and 50 are typically optimal for lower values of $K$, whereas for larger $K$ they generally range between 10 and 25. This suggests that, as the heterogeneity in feature relevance increases, a smaller contribution of the entropy regularization term in Eq. \eqref{eq:FWJM} is required to achieve an accurate recovery of the true feature importance structure.
%

\begin{figure}[!htbp] 
    \centering

    \begin{subfigure}{.48\linewidth}
        \centering
        \caption{$K=2$, $\alpha=0\%$}
        \includegraphics[width=\linewidth]{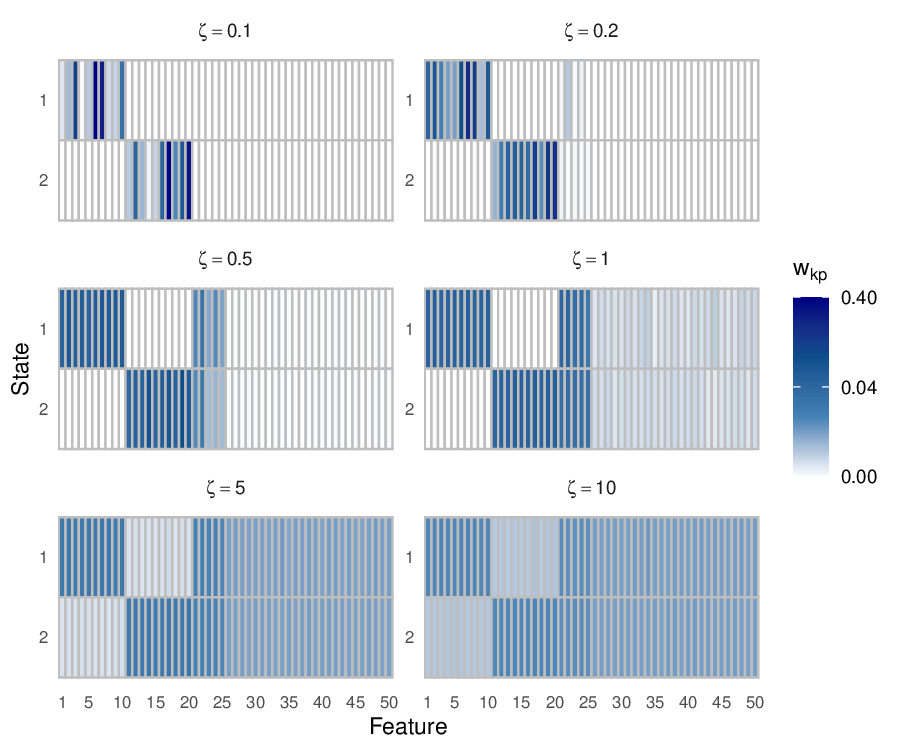}
    \end{subfigure}
    \begin{subfigure}{.48\linewidth}
        \centering
        \caption{$K=2$, $\alpha=5\%$}
        \includegraphics[width=\linewidth]{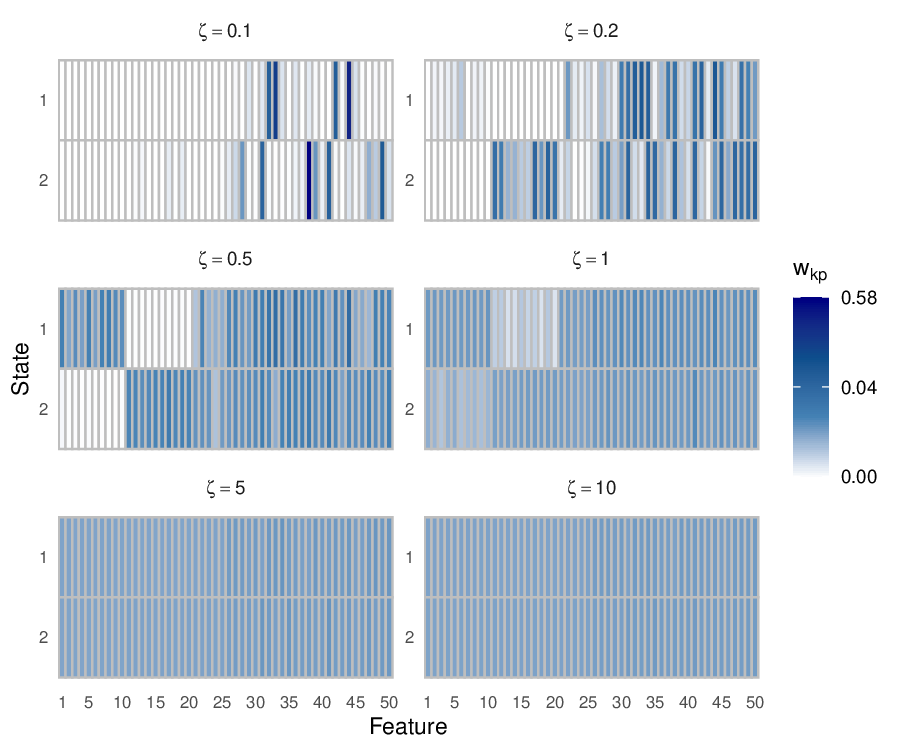}
    \end{subfigure}

    \begin{subfigure}{.48\linewidth}
        \centering
        \caption{$K=3$, $\alpha=0\%$}
        \includegraphics[width=\linewidth]{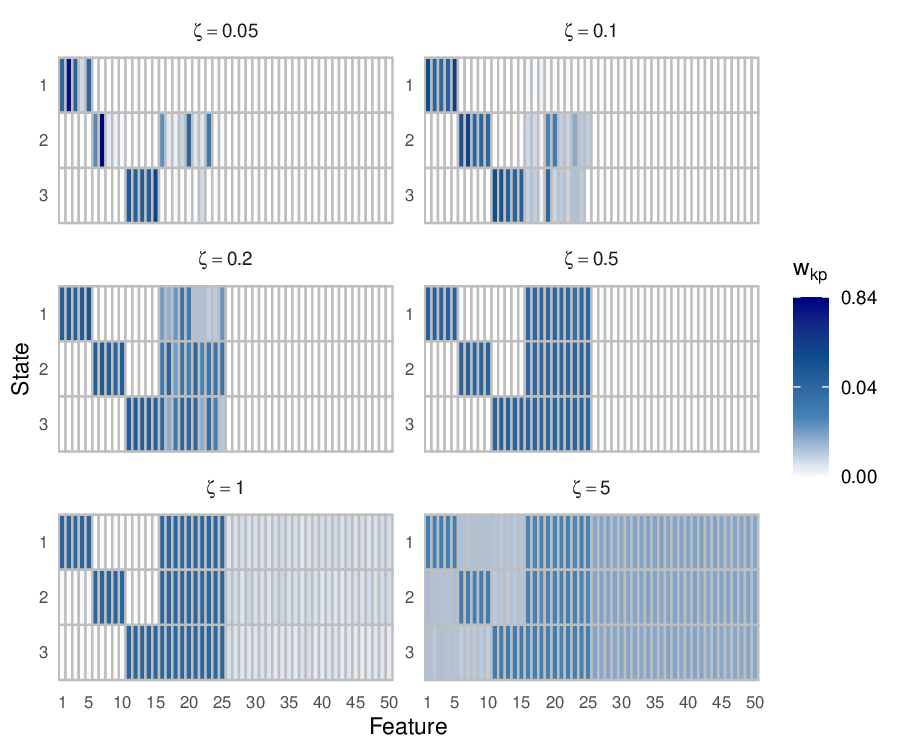}
    \end{subfigure}
    \begin{subfigure}{.48\linewidth}
        \centering
        \caption{$K=3$, $\alpha=5\%$}
        \includegraphics[width=\linewidth]{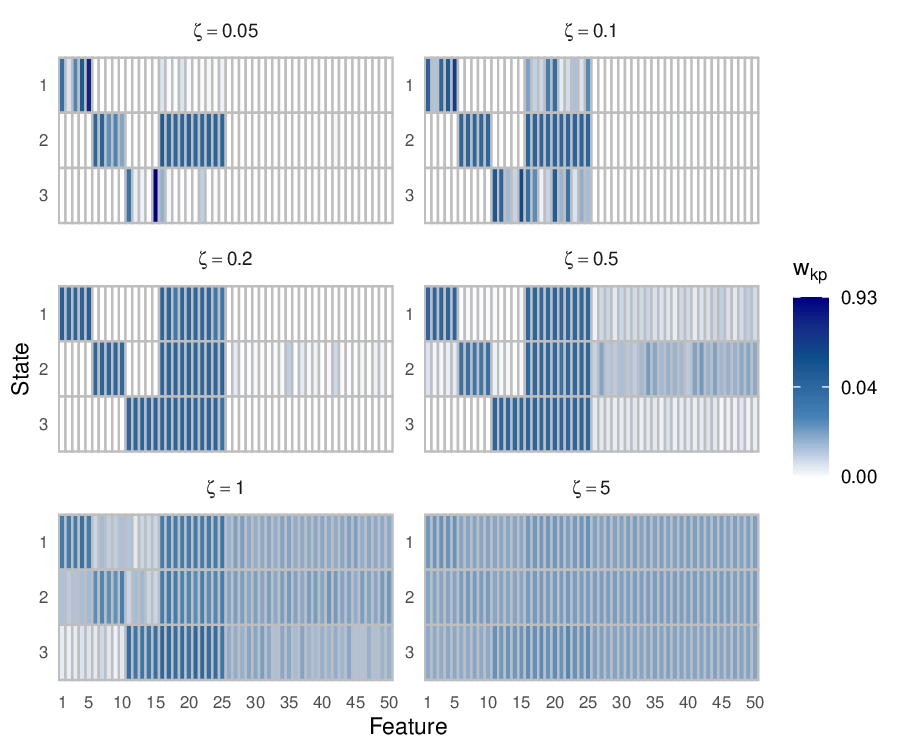}
    \end{subfigure}

    \begin{subfigure}{.48\linewidth}
        \centering
        \caption{$K=4$, $\alpha=0\%$}
        \includegraphics[width=\linewidth]{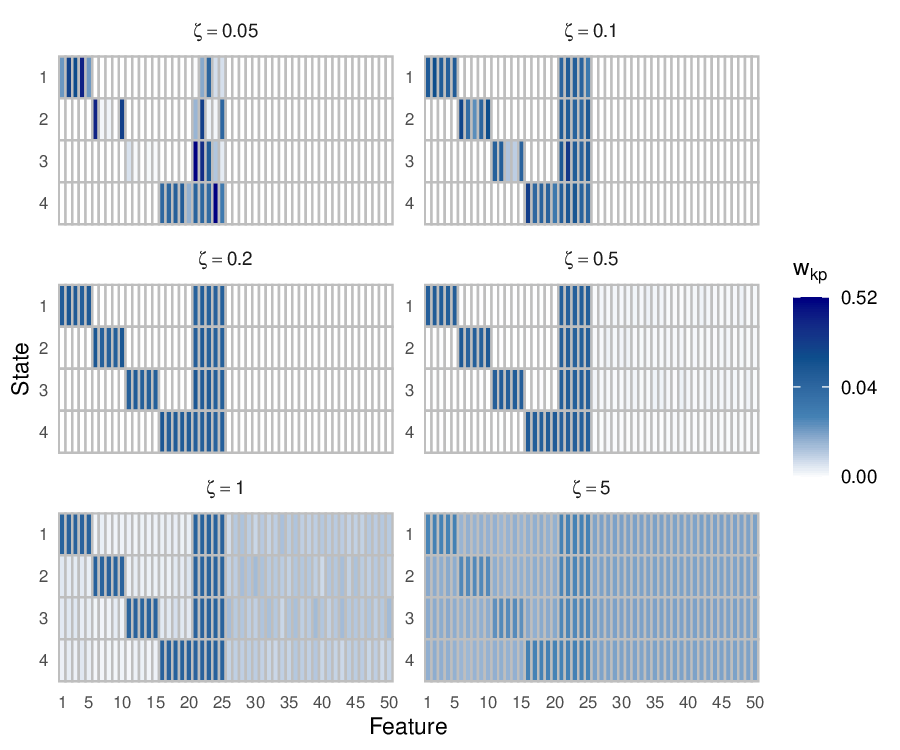}
    \end{subfigure}
    \begin{subfigure}{.48\linewidth}
        \centering
        \caption{$K=4$, $\alpha=5\%$}
        \includegraphics[width=\linewidth]{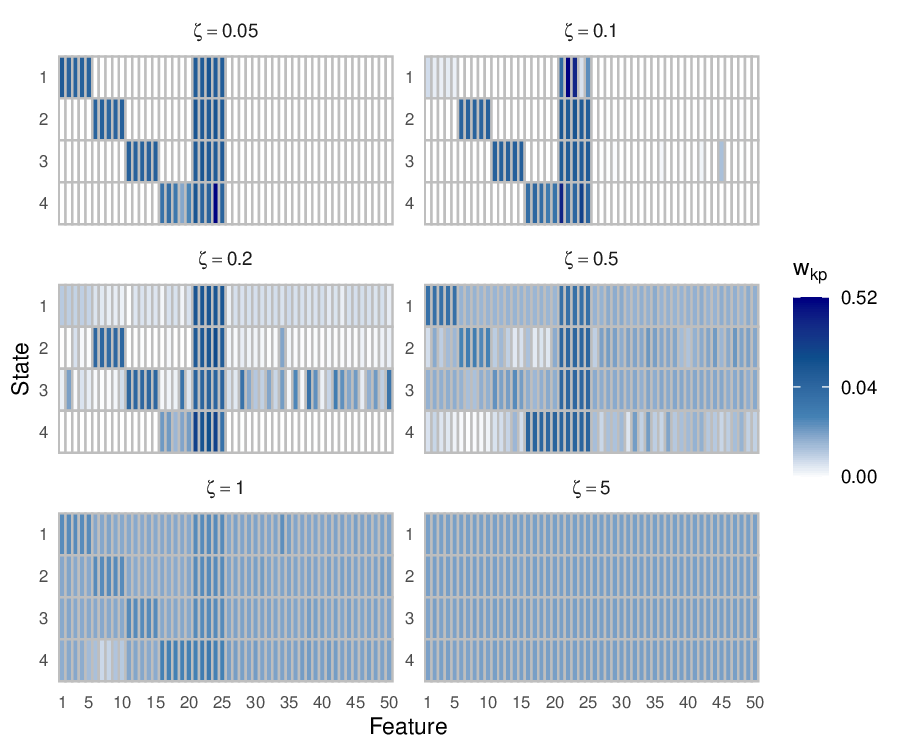}
    \end{subfigure}
    
    \caption{
    Normalized median weight matrices for FWJM under scenario B, with $0\%$ and $5\%$ contamination, for $K=2, K=3$, and $K=4$ latent states.
    }
    \label{fig:est_W_scenB_out}
\end{figure}

Figure \ref{fig:est_W_scenB_out} shows analogous results for scenario B, with the no-contamination setting shown on the left and the 5\% contamination setting on the right.
Optimal values of $\zeta$ lie in a smaller range, typically between $0.1$ and $1$, reflecting the higher level of sparsity in this scenario.
In the absence of contamination, the model robustly recovers the true feature-state structure, while performance deteriorates in the contaminated setting, mostly because this scenario is characterized by a small sample size. Nevertheless, for $K=3$ and $K=4$ the model is still able to recover the main feature-state structure. 
In contrast, when $K=2$, where inter-state feature heterogeneity is less pronounced, the model mainly identifies features that are strongly state-specific, while features shared across states and noisy features tend to receive similar weights.

Results for scenario C indicate that FWJM accurately recovers the underlying sparse state-feature structure across all values of $K$, even in the presence of contamination. 
In contrast, scenario D represents the most challenging setting due to the combination of small $T$ and high $P$, leading to a noticeable deterioration in performance  under contamination. Nevertheless, for sufficiently small values of $\zeta$ and in the absence of outliers, the model is still able to recover the main state-specific feature structure. 
Further details for these two scenarios are reported in Section~S1 of the Supplementary Material.

\subsection{Sensitivity Analysis}

%
%

\begin{figure}[!htbp] 
    \centering
    \begin{subfigure}{.48\linewidth}
        \centering
        \caption{$K=2$, $\alpha=0\%$}
        \includegraphics[width=\linewidth]{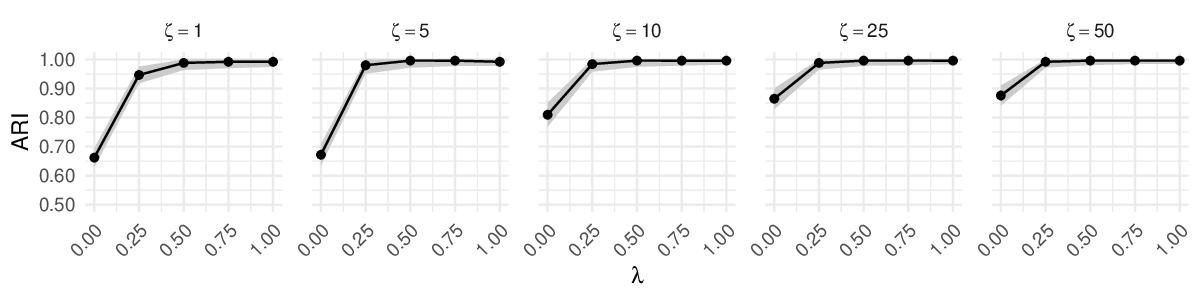}
        \includegraphics[width=\linewidth]{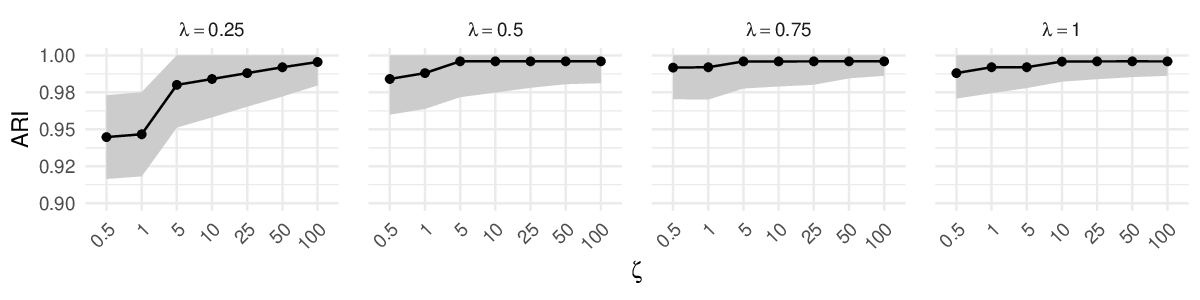}
        \label{fig:fwjm_A_out_K3}
    \end{subfigure}
    \begin{subfigure}{.48\linewidth}
        \centering
        \caption{$K=2$, $\alpha=5\%$}
        \includegraphics[width=\linewidth]{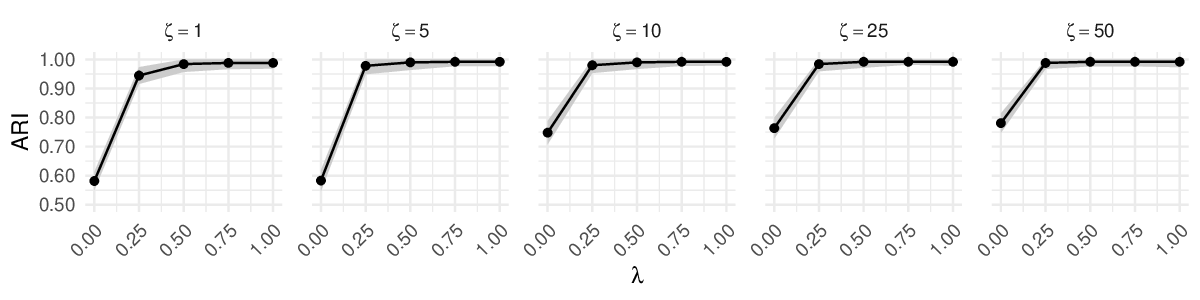}
        \includegraphics[width=\linewidth]{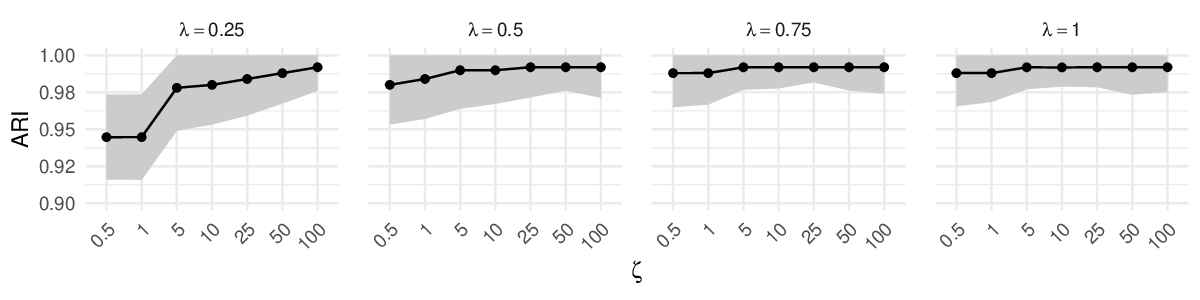}
        \label{fig:fwjm_A_out_K3}
    \end{subfigure}
    \hfill
    \begin{subfigure}{.48\linewidth}
        \centering
        \caption{$K=3$, $\alpha=0\%$}
        \includegraphics[width=\linewidth]{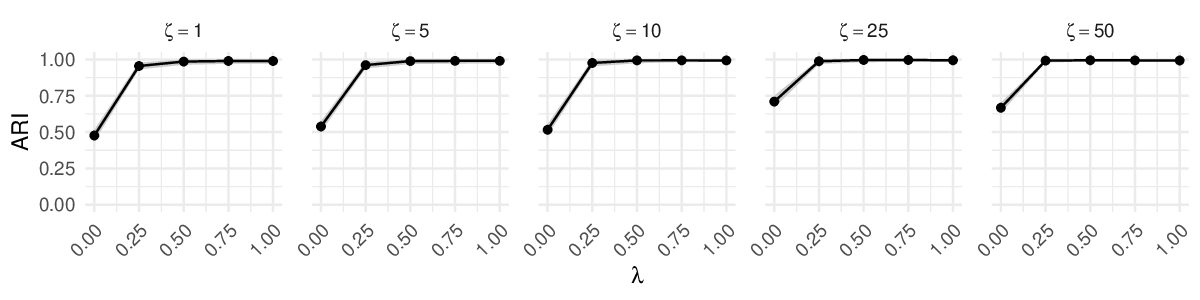}
        \includegraphics[width=\linewidth]{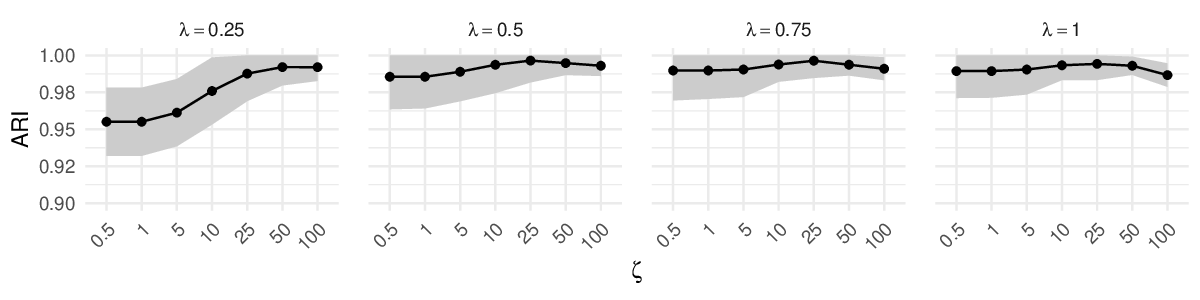}
        \label{fig:fwjm_A_out_K3}
    \end{subfigure}
    \hfill
     \begin{subfigure}{.48\linewidth}
        \centering
        \caption{$K=3$, $\alpha=5\%$}
        \includegraphics[width=\linewidth]{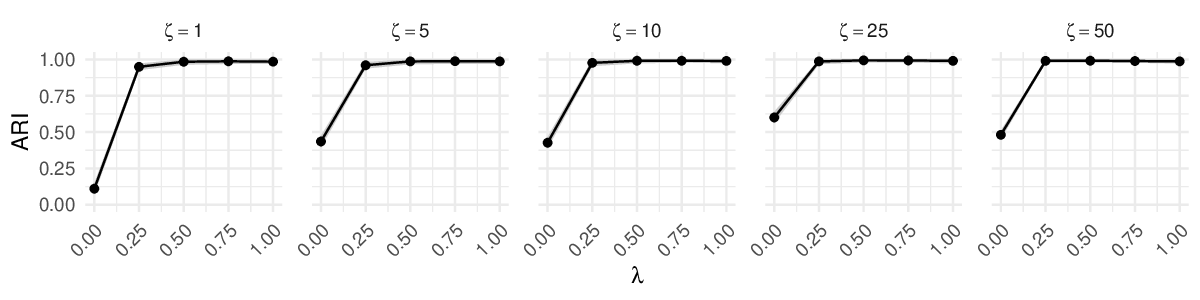}
        \includegraphics[width=\linewidth]{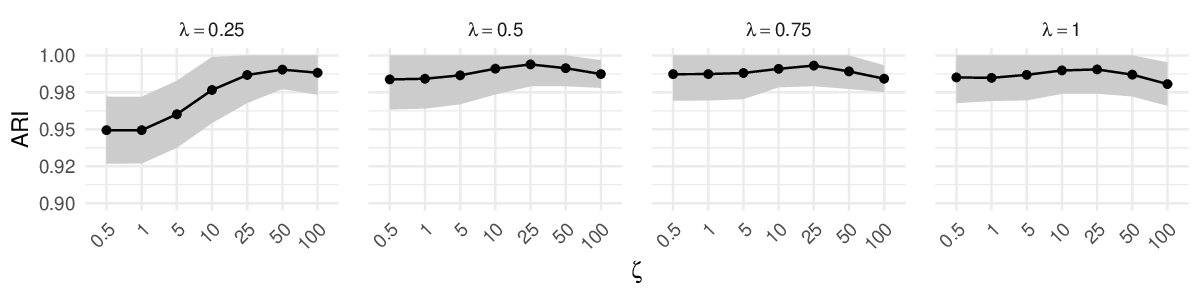}
        \label{fig:fwjm_A_out_K3}
    \end{subfigure}
    \hfill
    \begin{subfigure}{.48\linewidth}
        \centering
        \caption{$K=4$, $\alpha=0\%$}
        \includegraphics[width=\linewidth]{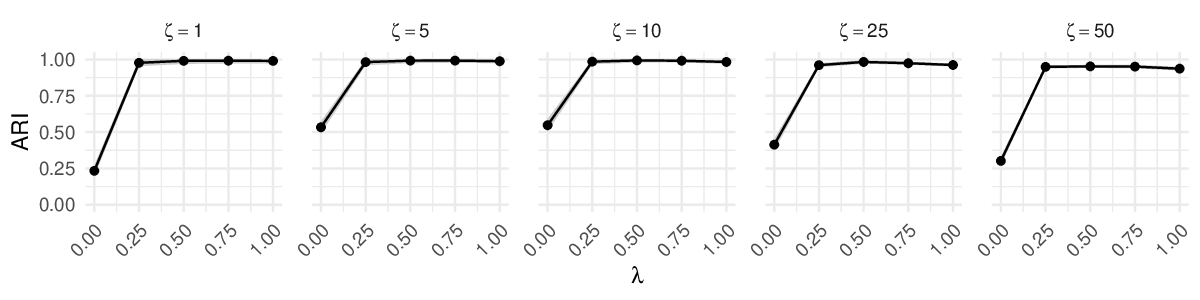}
        \includegraphics[width=\linewidth]{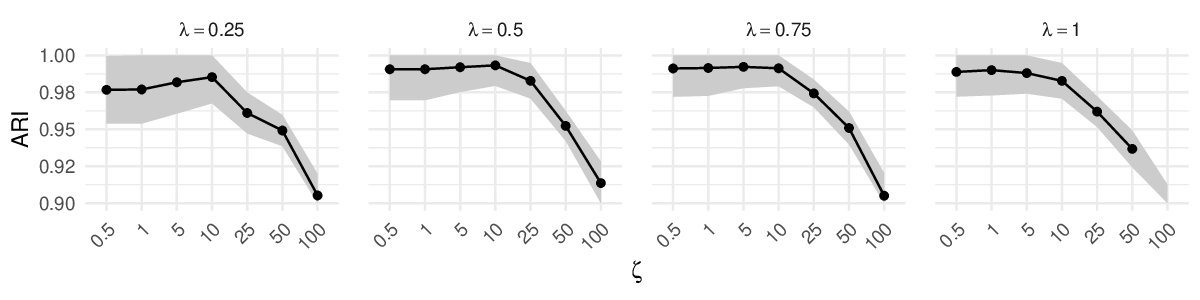}
        \label{fig:fwjm_A_out_K4}
    \end{subfigure}
    \hfill
    \begin{subfigure}{.48\linewidth}
        \centering
        \caption{$K=4$, $\alpha=5\%$}
        \includegraphics[width=\linewidth]{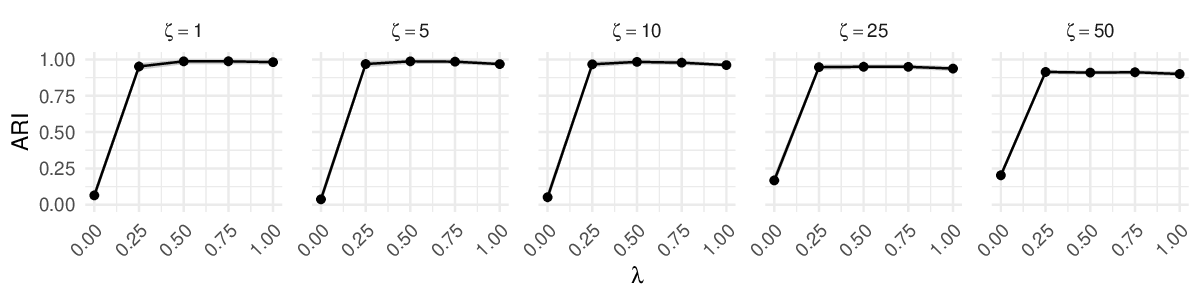}
        \includegraphics[width=\linewidth]{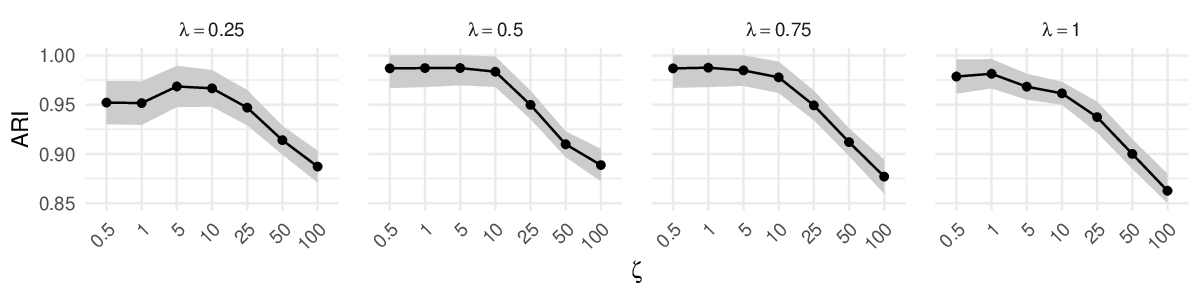}
        \label{fig:fwjm_A_out_K4}
    \end{subfigure}
    
    \caption{
    Sensitivity analysis for the feature-weighted jump model (FWJM) under scenario A, with $0\%$ and $5\%$ contamination, for $K=2, K=3$ and $K=4$ latent states. Each panel reports the median adjusted Rand index (ARI) as a function of the tuning parameters $\lambda$ and $\zeta$.
    Grey shaded areas denote 95\% confidence intervals.
    }
    \label{fig:sens_A}
\end{figure}

We now analyze the role of the hyperparameters $\zeta$ and $\lambda$ through a sensitivity analysis based on clustering accuracy.
Figure~\ref{fig:sens_A} reports ARI for varying values of $\lambda$ and $\zeta$ in scenario A.
The role of $\lambda$ is particularly evident, as introducing temporal regularization ($\lambda>0$) leads the ARI to increase close to one across all settings.
Clustering accuracy remains consistently high ($\mathrm{ARI} \approx 1$) over a broad range of $\zeta$ values when $K=2$ or $K=3$. In contrast, for $K=4$ the ARI decreases to approximately $0.9$ when $\zeta > 10$, confirming that when heterogeneity in the feature-state structure increases, the choice of $\zeta$ becomes more critical.
%

\begin{figure}[!htbp] 
\centering
\begin{subfigure}{.48\linewidth}
\centering
\caption{$K=2$, $\alpha=0\%$}
\includegraphics[width=\linewidth]{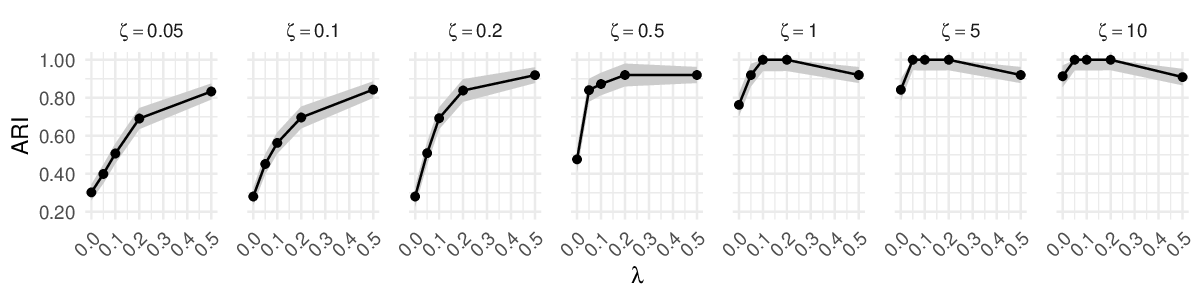}
\includegraphics[width=\linewidth]{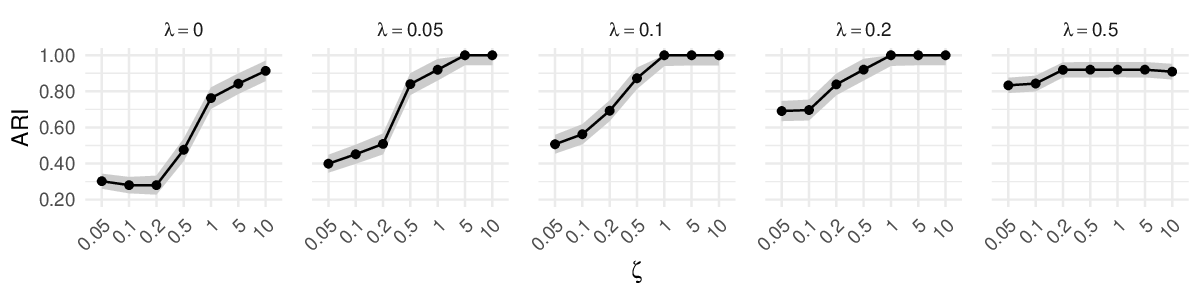}
\label{fig:fwjm_B_out_K3}
\end{subfigure}
\begin{subfigure}{.48\linewidth}
\centering
\caption{$K=2$, $\alpha=5\%$}
\includegraphics[width=\linewidth]{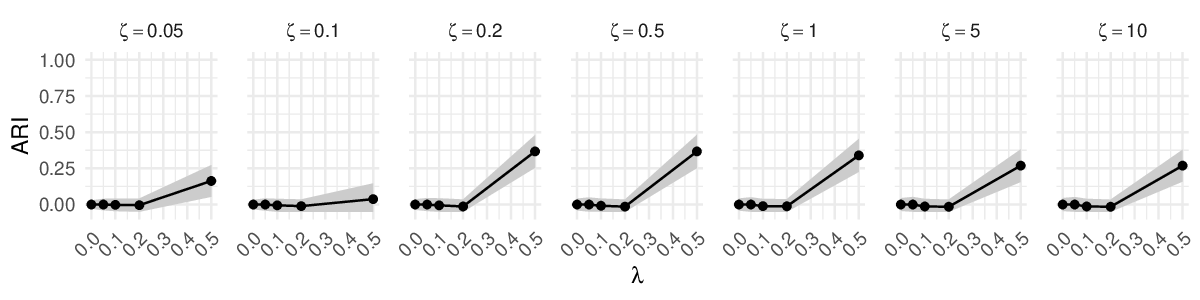}
\includegraphics[width=\linewidth]{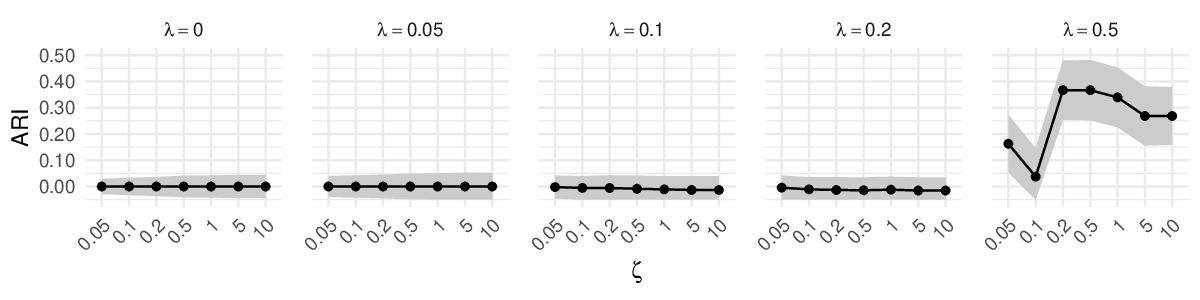}
\label{fig:fwjm_B_out_K3}
\end{subfigure}
\hfill
\begin{subfigure}{.48\linewidth}
\centering
\caption{$K=3$, $\alpha=0\%$}
\includegraphics[width=\linewidth]{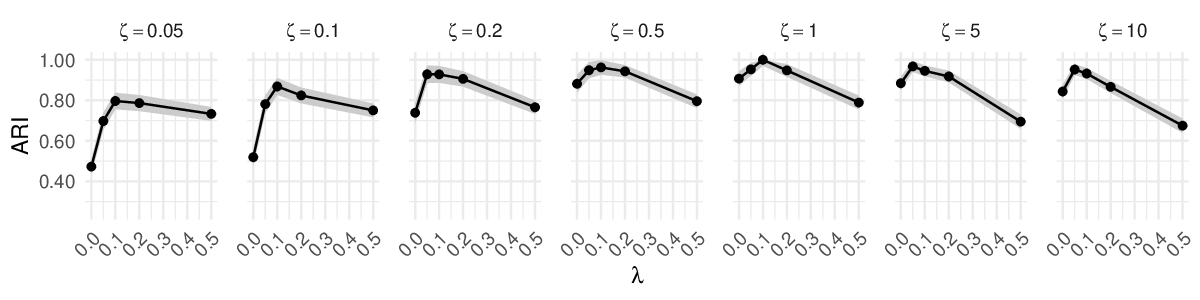}
\includegraphics[width=\linewidth]{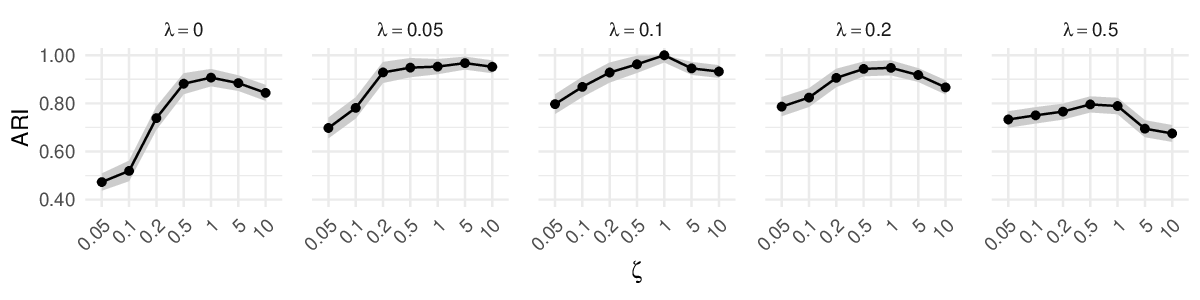}
\label{fig:fwjm_B_out_K3}
\end{subfigure}
\begin{subfigure}{.48\linewidth}
\centering
\caption{$K=3$, $\alpha=5\%$}
\includegraphics[width=\linewidth]{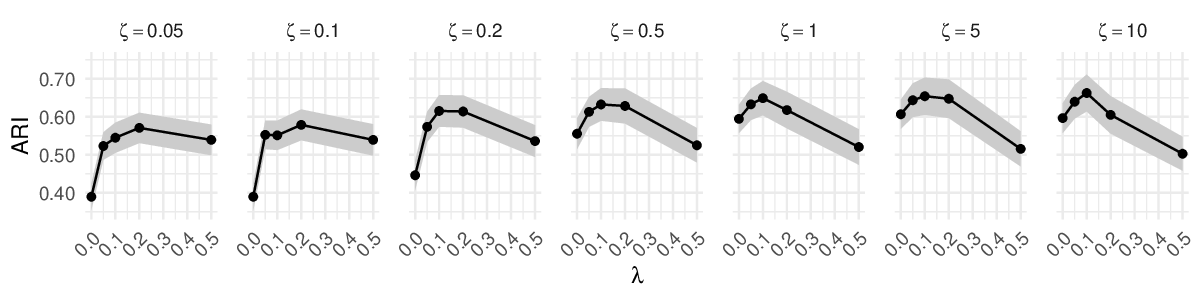}
\includegraphics[width=\linewidth]{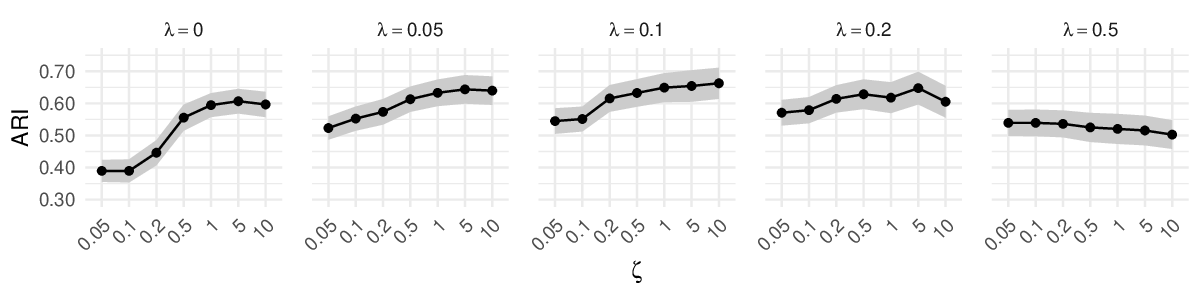}
\label{fig:fwjm_B_out_K3}
\end{subfigure}
\hfill
\begin{subfigure}{.48\linewidth}
\centering
\caption{$K=4$, $\alpha=0\%$}
\includegraphics[width=\linewidth]{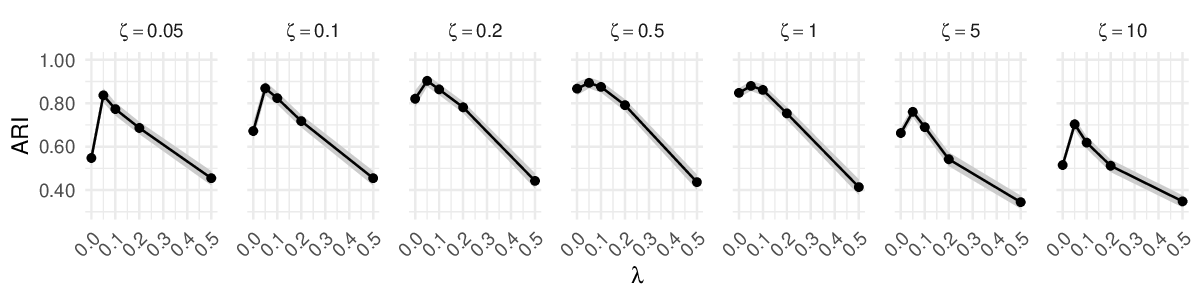}
\includegraphics[width=\linewidth]{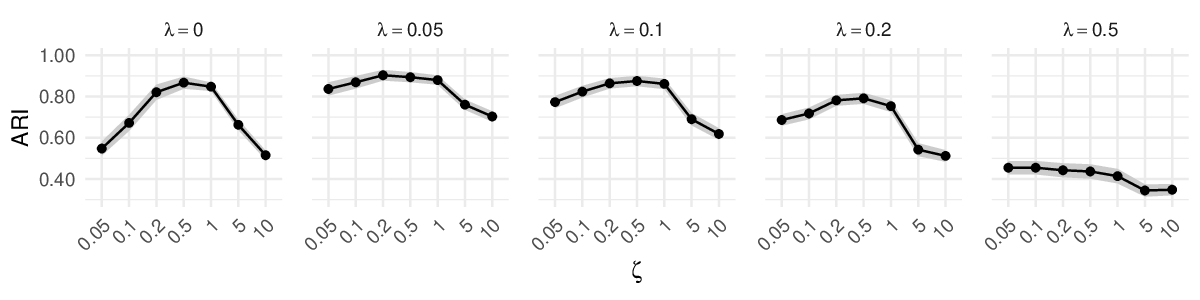}
\label{fig:fwjm_B_out_K4}
\end{subfigure}
\begin{subfigure}{.48\linewidth}
\centering
\caption{$K=4$, $\alpha=5\%$}
\includegraphics[width=\linewidth]{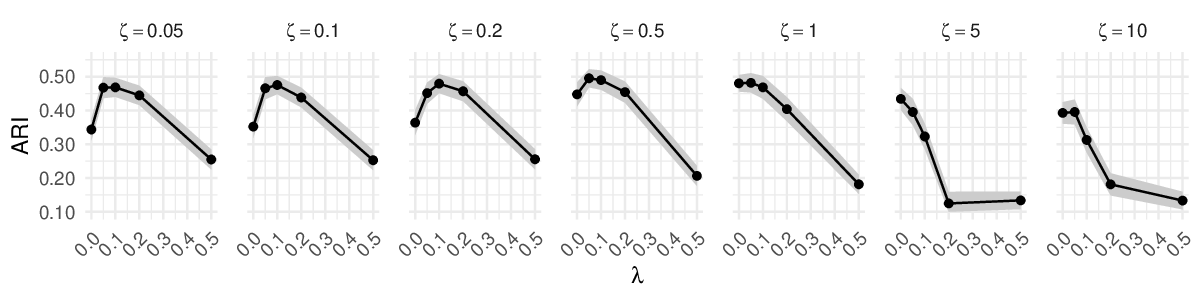}
\includegraphics[width=\linewidth]{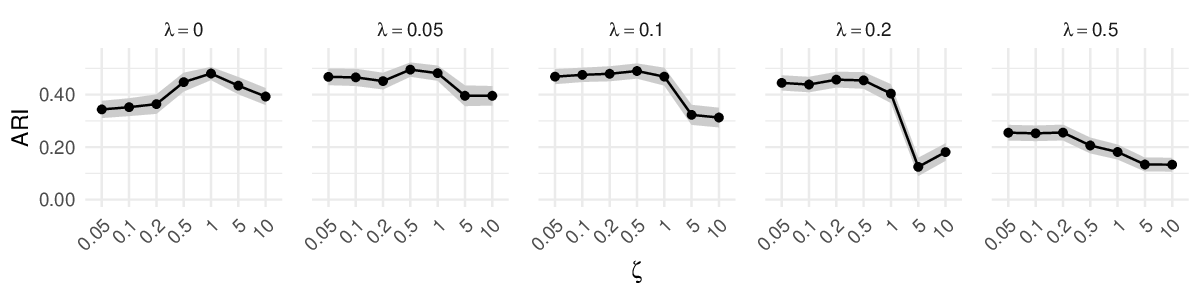}
\label{subfig:sens_B}
\end{subfigure}

\caption{
  Sensitivity analysis for the feature-weighted jump model (FWJM) under scenario B, with $0\%$ and $5\%$ contamination, for $K=2, K=3$ and $K=4$ latent states. Each panel reports the median adjusted Rand index (ARI) as a function of the tuning parameters $\lambda$ and $\zeta$.
  Grey shaded areas denote 95\% confidence intervals.
}
\label{fig:sens_B}
\end{figure}

Figure~\ref{fig:sens_B} reports the same sensitivity analysis for scenario B. In this setting, performance generally deteriorates as $\lambda$ increases when $K=3$ and $K=4$, reflecting the lower persistence of the latent states combined with the shorter time horizon, whereas the method is substantially more robust to $\lambda$ when $K=2$. 
Smaller values of $\zeta$, typically around $1$, generally achieve the highest ARI values, as weaker entropy regularization is preferable in this sparse high-noise setting. 
As expected, the introduction of outliers leads to a substantial deterioration in performance, particularly for $K=2$, where ARI values are often close to zero. In contrast, performance improves for $K=3$ and $K=4$, likely due to the greater inter-cluster heterogeneity, which makes the latent states more distinguishable.

Results for scenario C show that clustering performance is generally high and relatively robust to the choice of both tuning parameters across all values of $K$, even under contamination. 
In contrast, scenario D is substantially more challenging due to the combination of small $T$ and high $P$, resulting in satisfactory performance mainly in the absence of outliers. In this setting, good clustering performance is typically obtained only for relatively small values of both $\lambda$ and $\zeta$. 
Additional sensitivity analyses for the remaining scenarios are reported in Section~S1 of the Supplementary Material.

Overall, the sensitivity analysis highlights a clear interaction between the optimal tuning parameters and the dimensional and temporal structure of the data.
The persistence parameter $\lambda$ is mainly driven by the length of the time series. When $T$ is large, 
values of $\lambda>0$ improve clustering accuracy. 
%
The entropy parameter $\zeta$ is instead closely related to the overall dataset dimension and sparsity structure. When $P$ is small and most features are informative, as in scenario A, 
larger values of $\zeta$ are beneficial, with optimal values typically ranging between $10$ and $50$. 
When $P$ is large and many variables are noisy, as in scenarios B and C, smaller or moderate values of $\zeta$ are generally required to avoid assigning excessive weight to irrelevant features. 
%
Scenario D combines the most unfavorable conditions, namely small $T$ and large $P$. In such cases, both hyperparameters should be kept relatively small.

\section{Real Data Applications}
\label{sec:applications}

In this section, we illustrate our proposal with two different applications in different areas. In the first case, $T>>P$; while in the second case $T$ and $P$ are of the same order of magnitude. In real data, we need to select different hyperparameters: to do so, we specify a grid of values and estimate the model separately on each possible combination. For each combination of hyperparameters, we then compute the Silhouette width for each $t=1,\ldots,T$. We select the combination of hyperparameters corresponding to the largest {\it median} silhouette width, where the median is preferred to the mean to curb the influence of outliers. 

\subsection{Kosovo killings in the period 1998-2000}

The Kosovo conflict represents one of the most extensively documented conflicts in recent history. During the escalation of hostilities between Serbian forces and the Kosovo Liberation Army, the civilian population was exposed to brutal violence, forced displacement, and large-scale destruction of infrastructure. Persecutions, homicides, and forced displacements were carried out with the general intention of ethnic cleansing. 
Kosovo represents a unique real-world case study in the conflict literature: memorialisation efforts have led to
a rather accurate post-facto list of victims, recorded in the Kosovo Memory Book (\url{https://www.hlc-rdc.org/en/human-losses/kosovo-memory-book-ljudski-gubici/kosovo-memory-book/}).
We focus here on daily counts of civilian males and civilian females, and military (regardless of gender) killed in the period between 01/01/1998 and 16/12/2000. A total of 11536 conflict-related deaths (7050 civilian males, 1487 civilian females, and 2999 military) 
are reported in the Kosovo Memory Book. Most of the homicides occur in the Spring of 1999, as will be seen below. 

The resulting data set therefore contains daily counts of conflict-related deaths, with $T=1081$ days and $P=3$. The maximum observed count is 373 for civilian males, 88 for females, and 55 for the military. There are 448 days with no killings for any of the three categories in the period. 

In order to perform model selection, we let 
$\lambda=\{0,.25,.5,.75,.1,1.25\}$ $\zeta=\{.1,1,5,10,50,100,200,500\}$ and $K=\{2,3,4,5\}$. 
The median Silhouette width is usually very good, with 28.5\% of combinations exceeding 90\% and half exceeding 84\%. The maximum of 96.5\% is achieved in $K=2$, $\zeta=100$, and $\lambda=1$. 
A preference of the criterion for $K=2$ could be expected due to the large proportion of zeros, which creates a very strong separation between days without deaths and active war days. We will also show results related to the preferred combination when $K=3$, which is $\zeta=50$, and $\lambda=0$, and corresponds to a median Silhouette width of 94.1\%. 

First, we characterize the groups by reporting the average number of deaths for each variable for the times assigned to each group in Table \ref{KMBcentroids}.
 Clusters are clearly very well separated in both cases, with cluster-specific means that are positively associated, as expected. 

\begin{table}[!ht]
\centering
\begin{tabular}{lrrr}
  \hline
Cluster & Civ. M. & Civ. F. & Mil. \\ 
  \hline
  1 & 1.66 & 0.34 & 1.07 \\ 
  2 & 63.51 & 13.55 & 22.76 \\ 
  \hline 
  1 & 1.51 & 0.13 & 1.05 \\ 
  2 & 33.51 & 10.14 & 11.97 \\ 
  3 & 60.14 & 11.67 & 21.25 \\ 
   \hline
\end{tabular}
\caption{Average number of conflict-related deaths for civilian males and females, and military in Kosovo in the period 1998-2000. The results are stratified by the estimated cluster assignments in a two-cluster (upper panel) and three-cluster (lower panel) configuration.}
\label{KMBcentroids}
\end{table}

\begin{figure}[!htbp] 
\centering
\resizebox{\textwidth}{!}{%
\begin{tabular}{cc}
\begin{subfigure}[t]{0.48\linewidth}
    \centering
    \includegraphics[width=\linewidth]{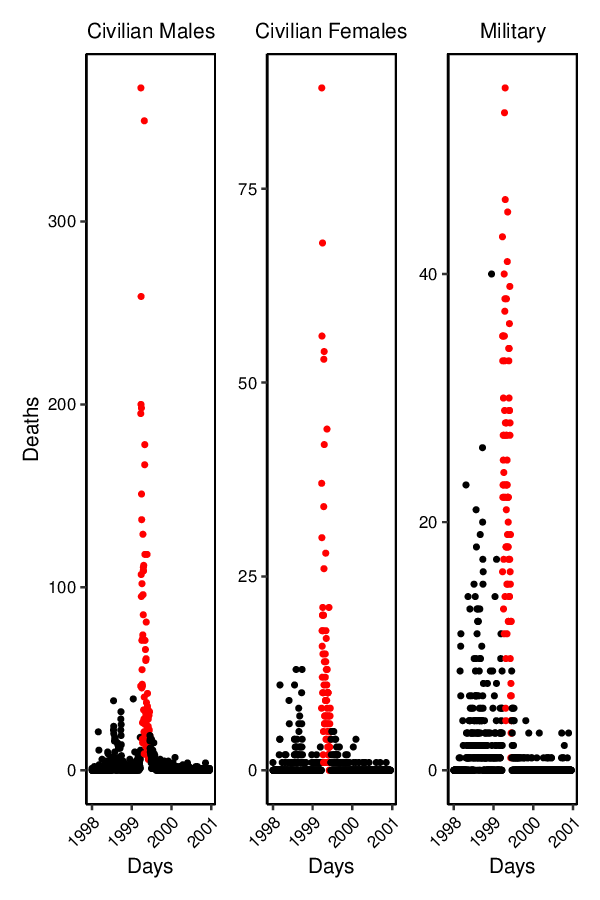}
    \caption{$K=2$}
\end{subfigure}
&
\begin{subfigure}[t]{0.48\linewidth}
    \centering
    \includegraphics[width=\linewidth]{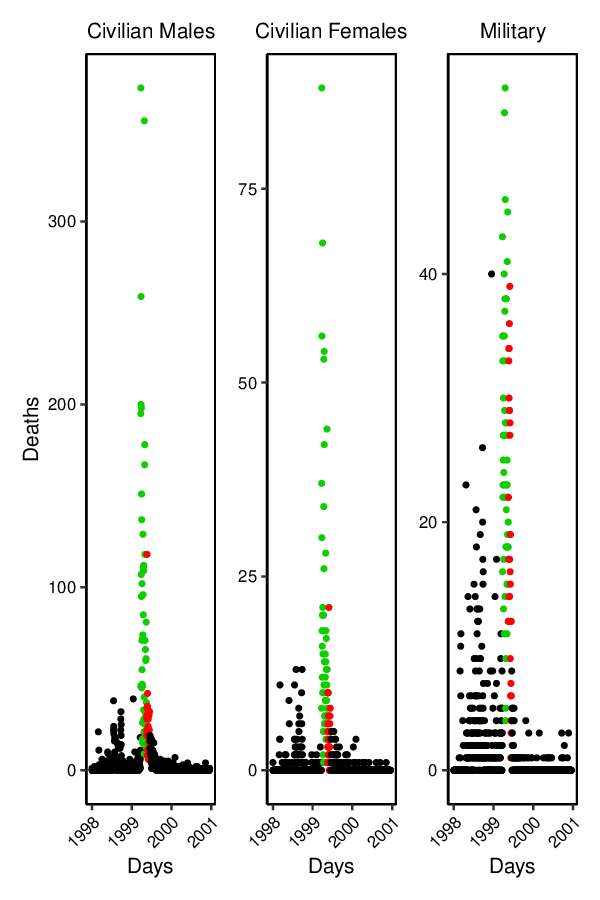}
    \caption{$K=3$}
\end{subfigure}

\end{tabular}
}
\caption{Time series of observed deaths in Kosovo in 1998-2000 by category. Points are colored differently according to the assigned cluster.}
\label{KMBts}
\end{figure}

In Figure \ref{KMBts} we plot the three observed time series, with different colors for each assigned cluster. 

The configuration with $K=2$ clearly identifies one period of high mortality risk for all indicators, while the remaining days are low risk,
corresponding to 
the brutal Spring of 1999. 
The configuration with $K=3$ further separates 
this cluster by assigning 
most of Spring 1999 to the new cluster 3; with some mixing between the two periods. 
The fact that days associated with the most violence are not uniformly labeled should not surprise, as the robust methodology is downweighting their influence on the final estimates. 

The estimated weight matrices are reported in Table \ref{KMBweights}. The column-wise heterogeneity observed supports our proposal, as the optimum is apparently far from the situation of cluster-homogeneous weights. 
For example, it can be seen that in the three-cluster configuration civilian females drive separation for cluster 1. 

\begin{table}[ht]
\centering
\begin{tabular}{lrrr}
  \hline
  Cluster & Civ. M. & Civ. F. & Mil. \\ 
  \hline
1 & 0.13 & 0.62 & 0.25 \\ 
  2 & 0.31 & 0.36 & 0.33 \\ 
  \hline
  1 & 0.03 & 0.88 & 0.09 \\ 
  2 & 0.25 & 0.45 & 0.30 \\ 
  3 & 0.29 & 0.39 & 0.32 \\ 
   \hline
\end{tabular}
\caption{Estimated weights for the proposed approach when $K=2$ (upper panel) and $K=3$ (lower panel) for the Kosovo killings data.}
\label{KMBweights}
\end{table}

We conclude this section with some evidence of the necessity of using Tukey's biweight loss for the data at hand, due to the presence of outlying days with a very large number of killings. In order to do so, we have re-estimated the model with a classical square loss, both with $K=2$ and $K=3$. When $K=2$ the weight matrix collapses and for the second group, the third variable (number of military deaths) receives 100\% of the weight. 
The same happens when $K=3$, and both cluster 2 and cluster 3 give 100\%  of the weight to military deaths. 
The centroids become closer to each other when $K=2$, although a good separation can still be observed. When $K=3$, the third cluster is associated with twice the average deaths of civilians than was reported above; leading to an over-estimation of the conflict burden on civilians for days assigned to clusters 2 and 3. 

\subsection{Macroeconomic indicators of Euro countries}

The area of macroeconomics, in our opinion, is strongly relevant for the methodology at hand. Multivariate time series of macroeconomic indicators are often available and, although they mostly evolve smoothly over time, they are also frequently characterized by structural breaks and jumps due to economic or political shocks, including wars, pandemics, market crashes, and terrorist threats. Furthermore, outliers are often present, and data analysis should take that into account by means of robust methods. 

We illustrate the use of our approach in macroeconomics with three indicators (household consumption to GDP, imports to GDP, and exports to GDP) for twelve European countries (Austria, Belgium, Estonia, Finland, France, Greece, Ireland, Italy, Malta, Netherlands, Portugal, and Spain), from 1949 to 2024, included. 

Hence, in this example, $T=49$ and $P=12*3=36$. 

For model selection, we compare all possible combinations of $\zeta=\{.05,.1,.2,.5,1,5,10\}$, $\lambda=\{0,.25,.5,.75,1\}$ and $K=\{2,3,4,5\}$. 

The maximum median Silhouette width of 53.5\% corresponds to $K=2$, $\zeta=1$, and $\lambda=.25$. 
A little more sensitivity to the hyperparameters than in the previous example could be expected, and this is indeed seen in the present data. This is due both to a less strong separation between groups (although a Silhouette above 50\% is still generally interpreted as a satisfactory cluster representation) and to the high dimensionality of the data. Only 34 (25\%) hyperparameter combinations correspond to a median silhouette width greater than 50\%; and 10 (7.1\%) are less than 25\%. 

For the selected parameters, fifteen of the 36 variables receive near zero weight for one cluster;  where all consumption measurements are essentially discarded; together with exports and imports of Estonia, which therefore has zero weight for the first cluster. 
Consumption of Italy, Malta, the Netherlands and Portugal have approximately zero weight also for the second cluster;  together with exports and imports of Greece and Malta. 
The weights are also very heterogeneous when comparing the two groups, with a median relative difference of 24\%.

\section{Conclusions}
\label{sec:conclusions}

In this work, we proposed a flexible framework for state-dependent variable weighting in robust dynamical clustering of multivariate time series. 
Simulation studies show that the proposal is able to accurately recover both the latent state sequence and the relevant features, even in the presence of noise variables, outliers, and complex temporal patterns. 
In the empirical applications presented, the necessity of all novel features specific to FWJM is strongly supported, including robust dissimilarities, cluster-specific weights, and time-dependent smoothing of the underlying cluster assignments. 

We stress that our method is devoted to variable weighting, while exact sparsity (i.e., variable selection) is not achieved as in general all weights are positive. This is an explicit choice, which is shared with many other methods (such as \cite{friedman2004clustering}). While in our experience, as illustrated also in the simulation study and real data examples, irrelevant features are typically down-weighted to an almost zero weight; we stress that in unsupervised learning, variable selection should arise also from reasoning (linked with the scientific question at hand). 
A clear limitation of our method is that it does not scale well with large $T$, as the dissimilarity between all possible pairs of measurements assigned to each cluster must be calculated at each iteration. Future work may be devoted to the development of scalable algorithms for large $T$ and/or high-dimensional settings.

\bibliography{biblio}%

\clearpage

\renewcommand{\thesection}{S\arabic{section}}
\setcounter{section}{0}


{\huge \centering Supplementary Material}

\vspace{3mm}

\noindent
Section \ref{sec:addres} reports additional simulation results. Section \ref{sec:deriv} derives the closed-form formula for the state-conditional feature weights.

\section{Additional simulation results}
\label{sec:addres}

\begin{figure}[!htbp] 
    \centering

    \begin{subfigure}{.48\linewidth}
        \centering
        \caption{$K=2$, $\alpha=0\%$}
        \includegraphics[width=\linewidth]{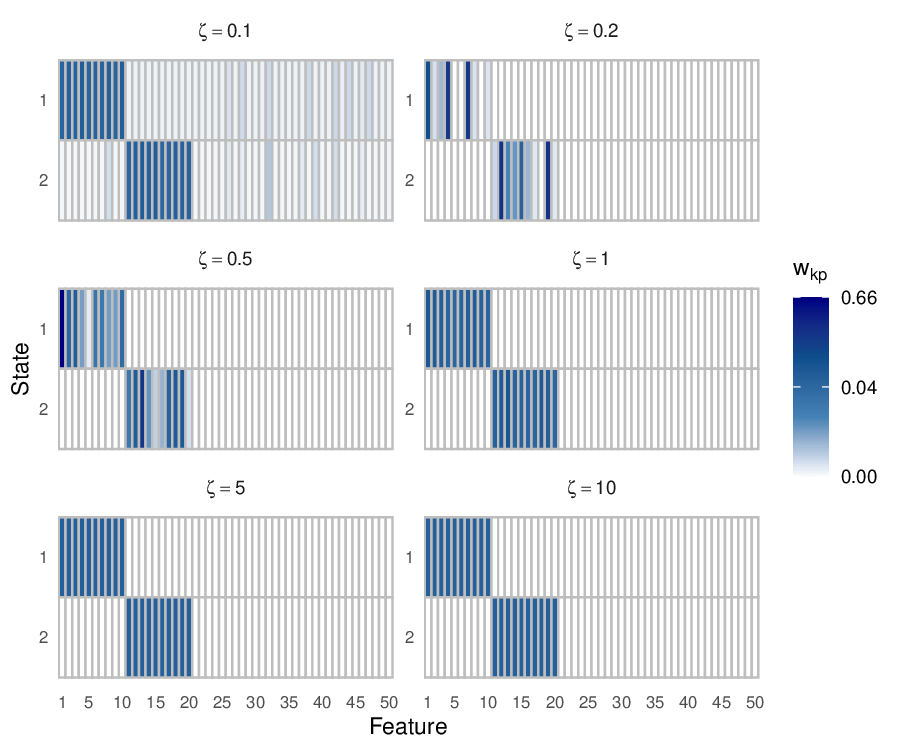}
    \end{subfigure}
    \begin{subfigure}{.48\linewidth}
        \centering
        \caption{$K=2$, $\alpha=5\%$}
        \includegraphics[width=\linewidth]{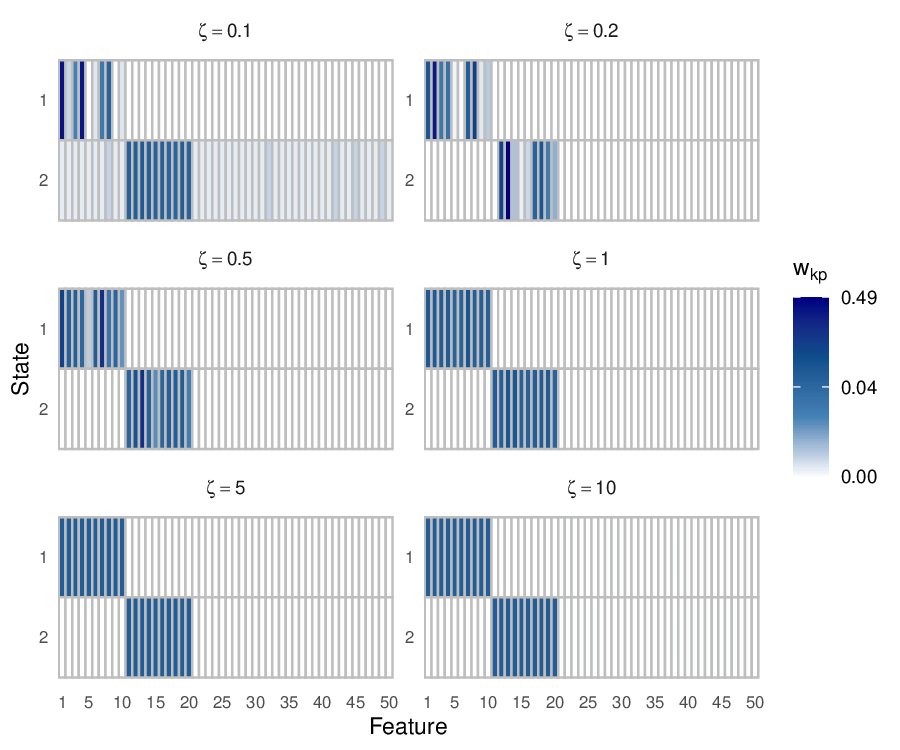}
    \end{subfigure}

    \begin{subfigure}{.48\linewidth}
        \centering
        \caption{$K=3$, $\alpha=0\%$}
        \includegraphics[width=\linewidth]{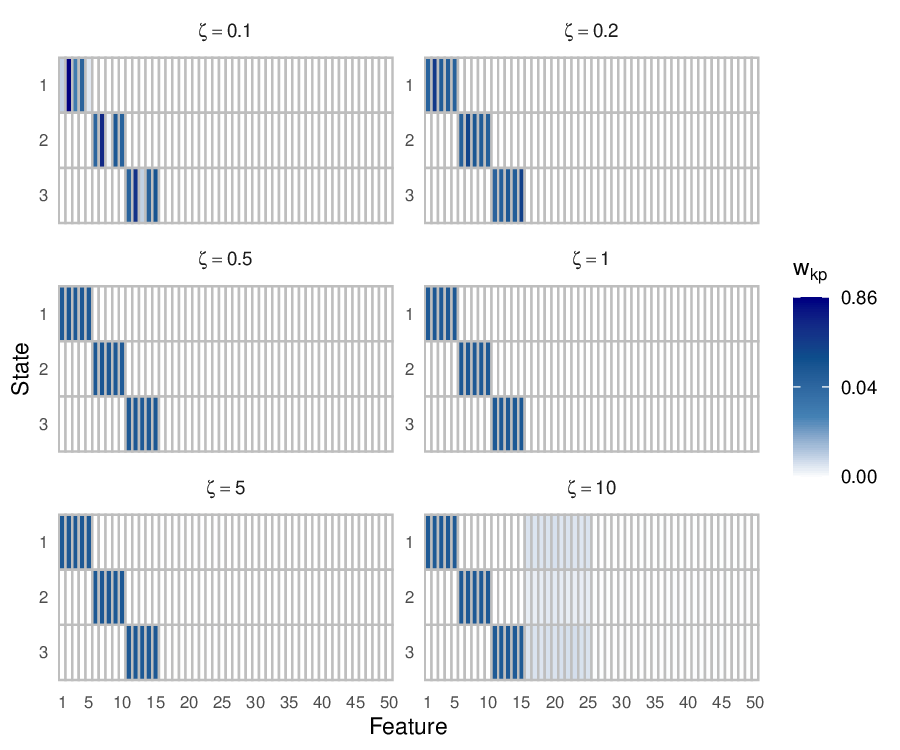}
    \end{subfigure}
    \begin{subfigure}{.48\linewidth}
        \centering
        \caption{$K=3$, $\alpha=5\%$}
        \includegraphics[width=\linewidth]{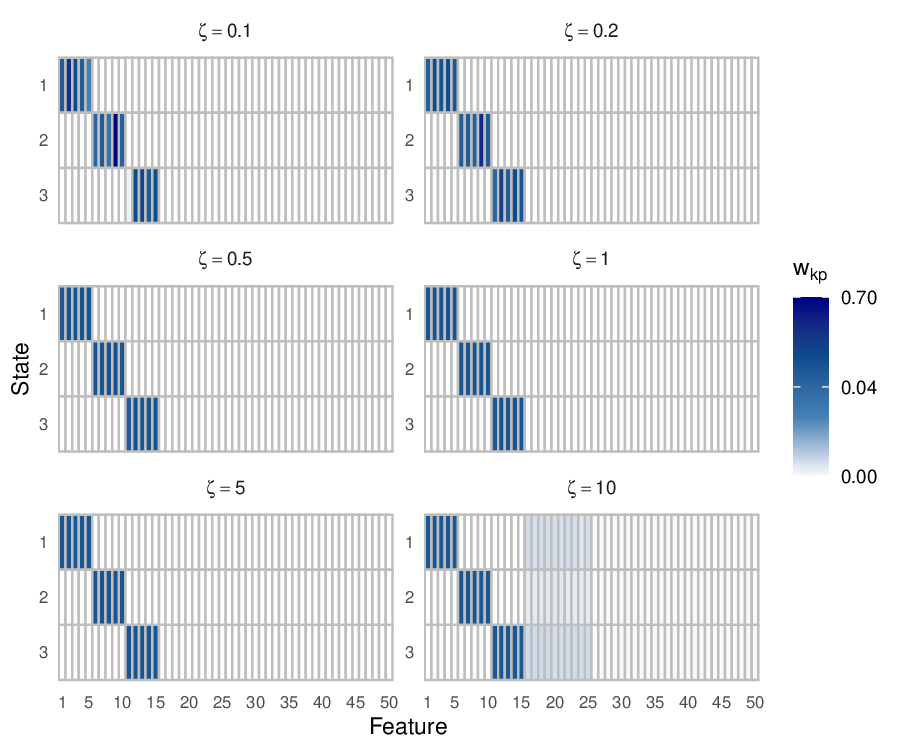}
    \end{subfigure}

    \begin{subfigure}{.48\linewidth}
        \centering
        \caption{$K=4$, $\alpha=0\%$}
        \includegraphics[width=\linewidth]{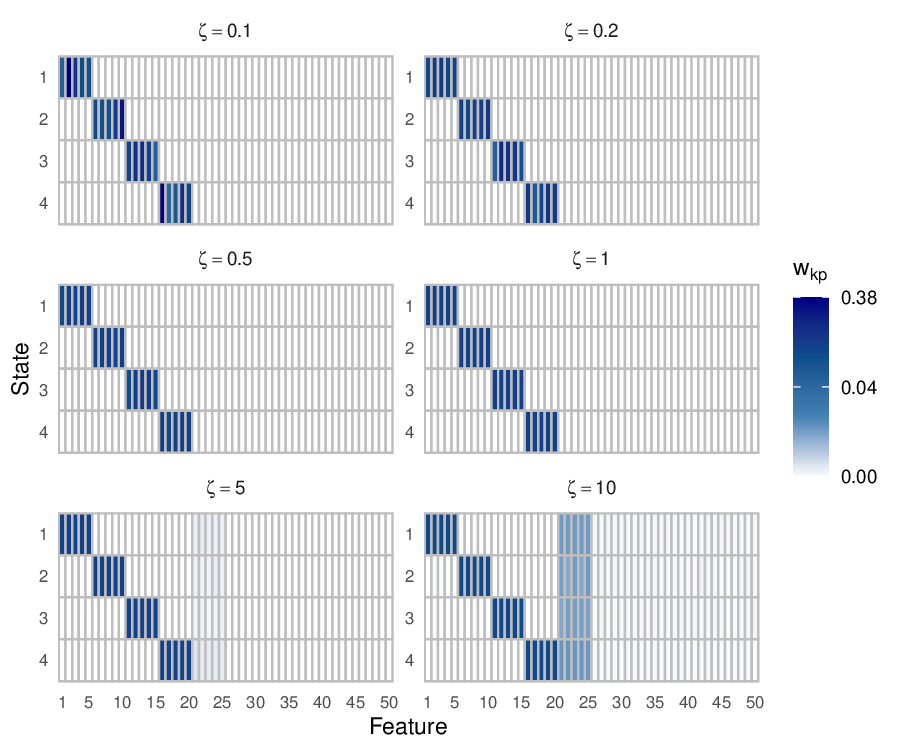}
    \end{subfigure}
    \begin{subfigure}{.48\linewidth}
        \centering
        \caption{$K=4$, $\alpha=5\%$}
        \includegraphics[width=\linewidth]{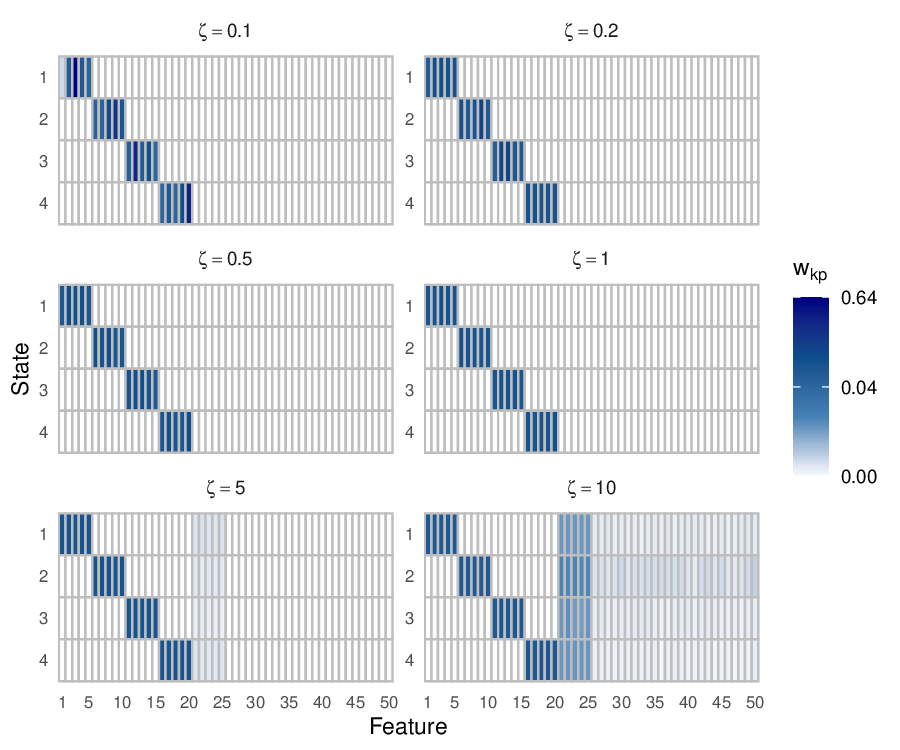}
    \end{subfigure}
    
    \caption{
    Normalized median weight matrices for FWJM under Scenario C, with $0\%$ and $5\%$ contamination, for $K=2, K=3$, and $K=4$ latent states.
    }
    \label{fig:est_W_scenC_out}
\end{figure}

Figure \ref{fig:est_W_scenC_out} reports the results for Scenario C, which corresponds to the richest setting in terms of data availability. In this case, the model perfectly recovers the true feature-state structure for values of $K=3$ and $4$, for all contamination levels.
For $K=2$, however, the model does not identify the five features shared across the two states, assigning them weights close to zero. This likely reflects the increased selectivity induced by the larger sample size: as the state-specific signal is estimated more accurately, the model tends to concentrate the weights on features that are genuinely discriminative between states. 

\begin{figure}[!htbp] 
    \centering

    \begin{subfigure}{.48\linewidth}
        \centering
        \caption{$K=2$, $\alpha=0\%$}
        \includegraphics[width=\linewidth]{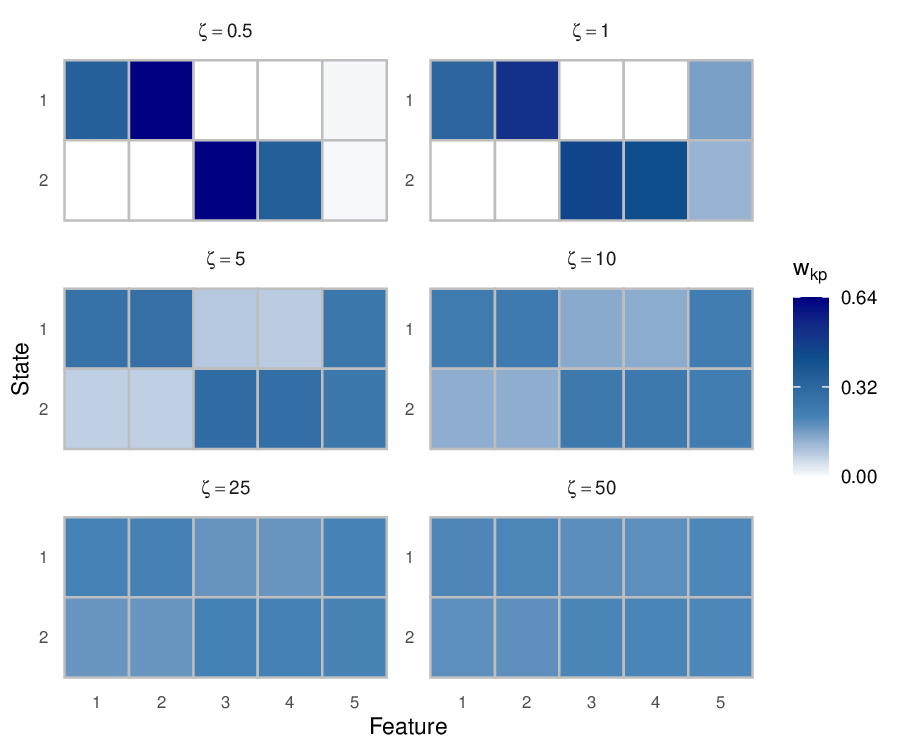}
    \end{subfigure}
    \begin{subfigure}{.48\linewidth}
        \centering
        \caption{$K=2$, $\alpha=5\%$}
        \includegraphics[width=\linewidth]{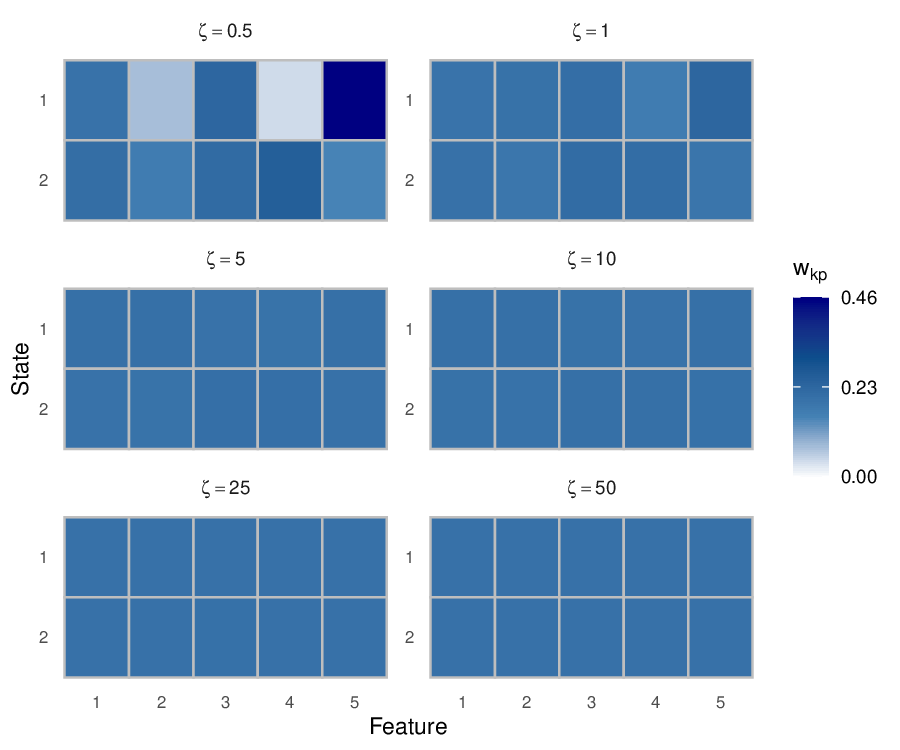}
    \end{subfigure}

    \begin{subfigure}{.48\linewidth}
        \centering
        \caption{$K=3$, $\alpha=0\%$}
        \includegraphics[width=\linewidth]{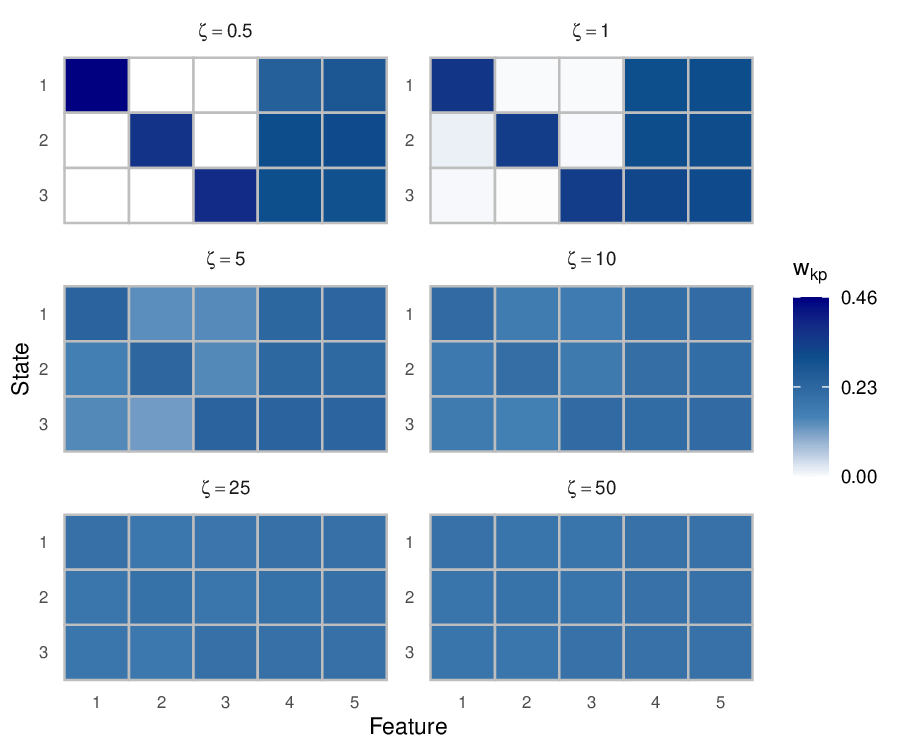}
    \end{subfigure}
    \begin{subfigure}{.48\linewidth}
        \centering
        \caption{$K=3$, $\alpha=5\%$}
        \includegraphics[width=\linewidth]{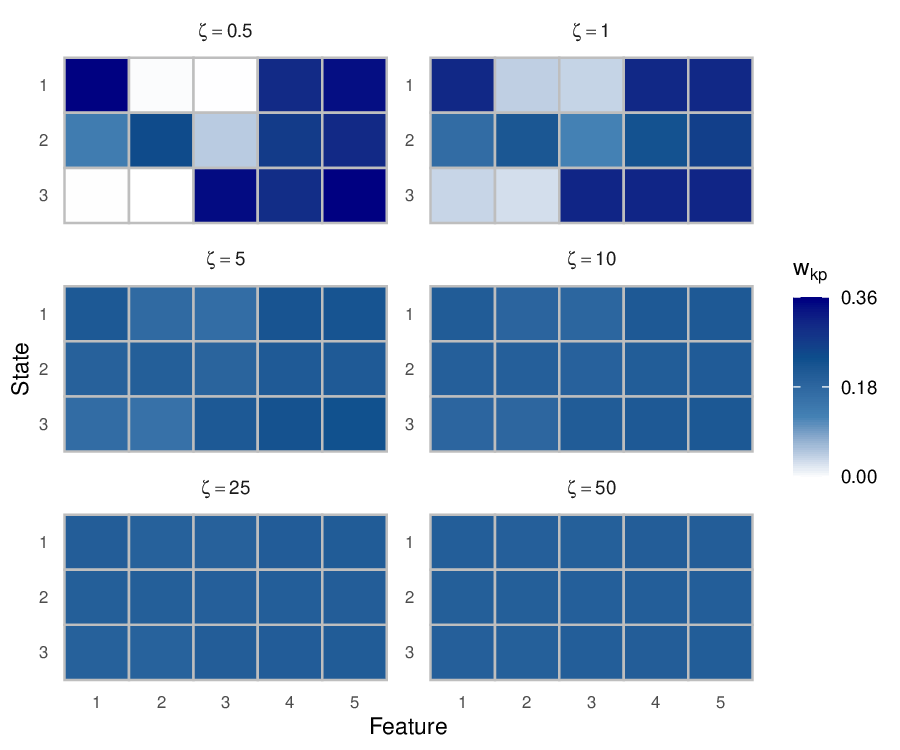}
    \end{subfigure}

    \begin{subfigure}{.48\linewidth}
        \centering
        \caption{$K=4$, $\alpha=0\%$}
        \includegraphics[width=\linewidth]{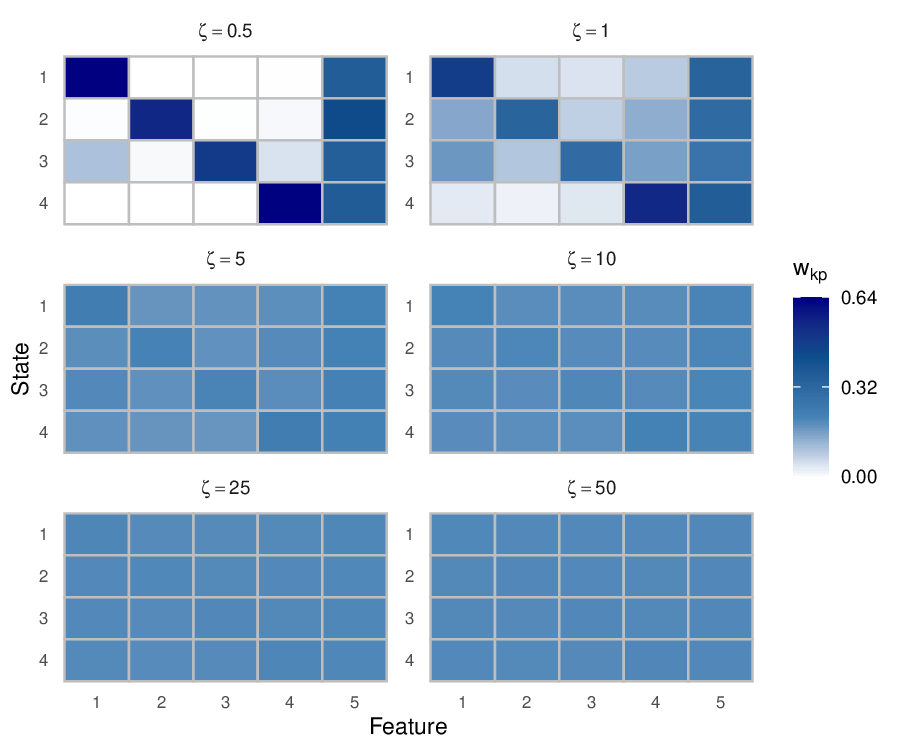}
    \end{subfigure}
    \begin{subfigure}{.48\linewidth}
        \centering
        \caption{$K=4$, $\alpha=5\%$}
        \includegraphics[width=\linewidth]{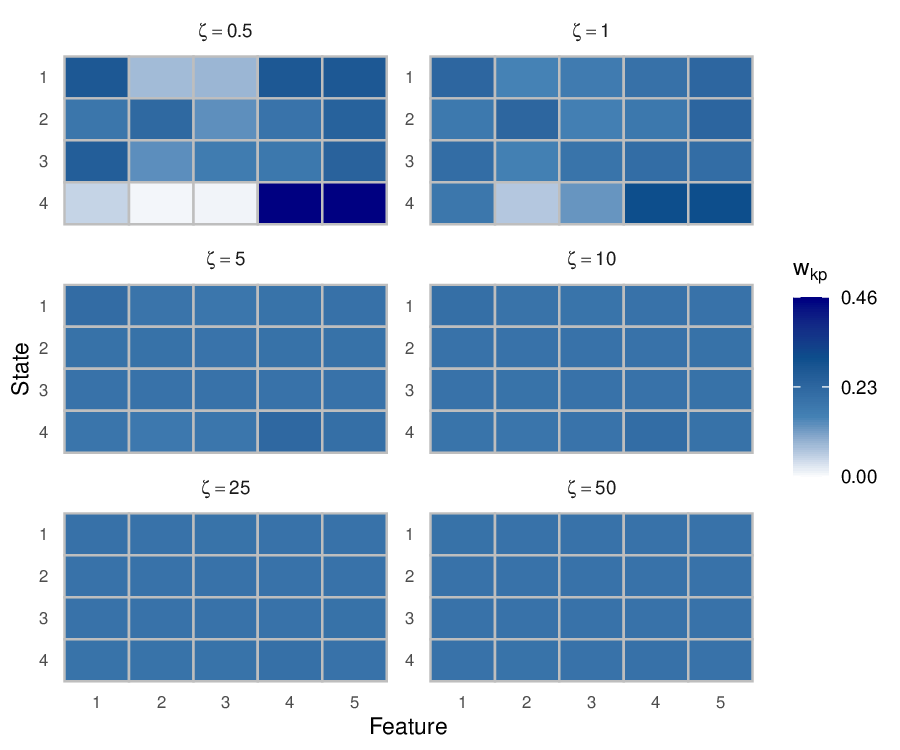}
    \end{subfigure}
    
    \caption{
    Normalized median weight matrices for FWJM under Scenario D, with $0\%$ and $5\%$ contamination, for $K=2, K=3$, and $K=4$ latent states.
    }
    \label{fig:est_W_scenD_out}
\end{figure}

Figure \ref{fig:est_W_scenD_out} reports the results for Scenario D, which represents the most challenging setting due to the combination of small $T$ and high $P$. Accordingly, performance deteriorates substantially, particularly in the presence of outliers.
Nevertheless, for small values of $\zeta$ and in the absence of contamination, the model is still able to recover the true feature-state structure for all values of $K$.

\begin{figure}[!htbp] 
\centering
\begin{subfigure}{.48\linewidth}
\centering
\caption{$K=2$, $\alpha=0\%$}
\includegraphics[width=\linewidth]{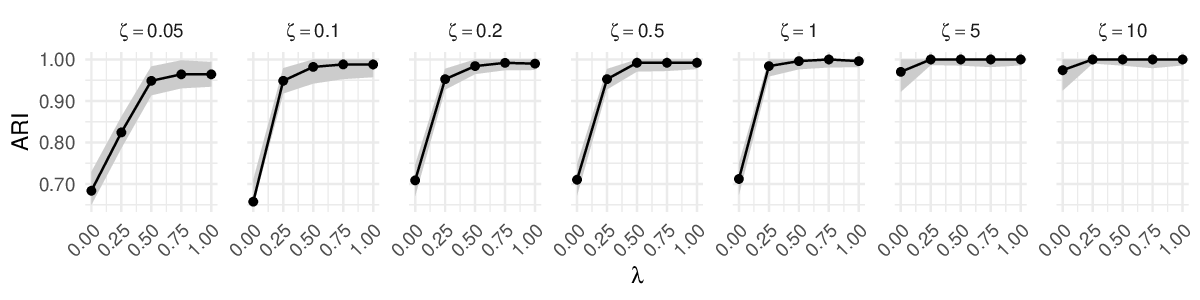}
\includegraphics[width=\linewidth]{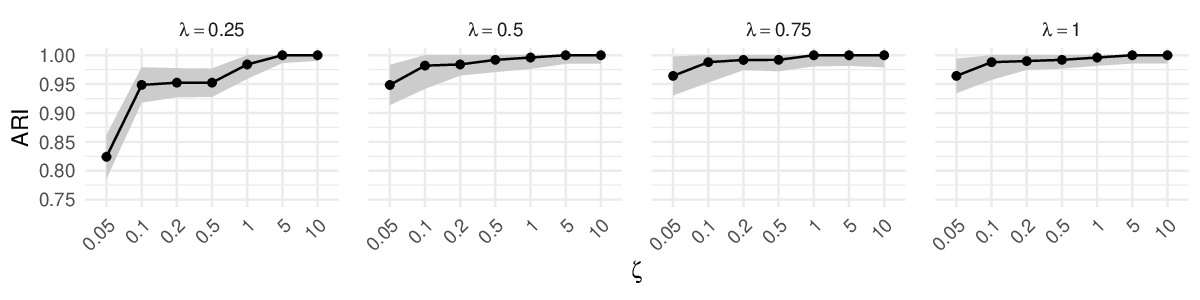}
\label{fig:fwjm_C_out_K3}
\end{subfigure}
\begin{subfigure}{.48\linewidth}
\centering
\caption{$K=2$, $\alpha=5\%$}
\includegraphics[width=\linewidth]{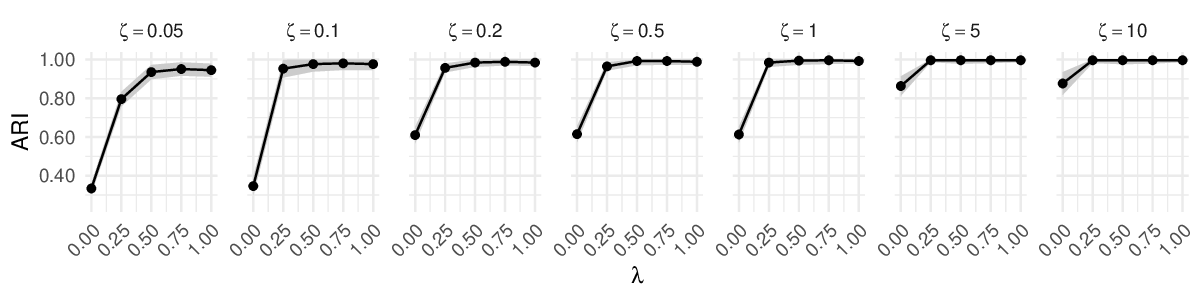}
\includegraphics[width=\linewidth]{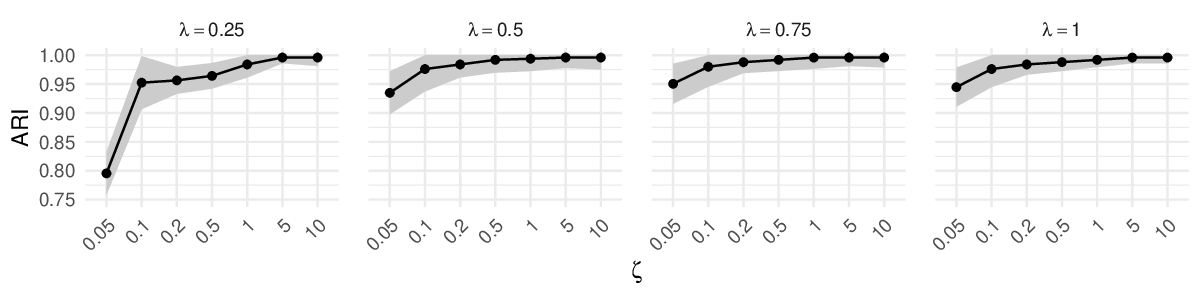}
\label{fig:fwjm_C_out_K3}
\end{subfigure}
\hfill
\begin{subfigure}{.48\linewidth}
\centering
\caption{$K=3$, $\alpha=0\%$}
\includegraphics[width=\linewidth]{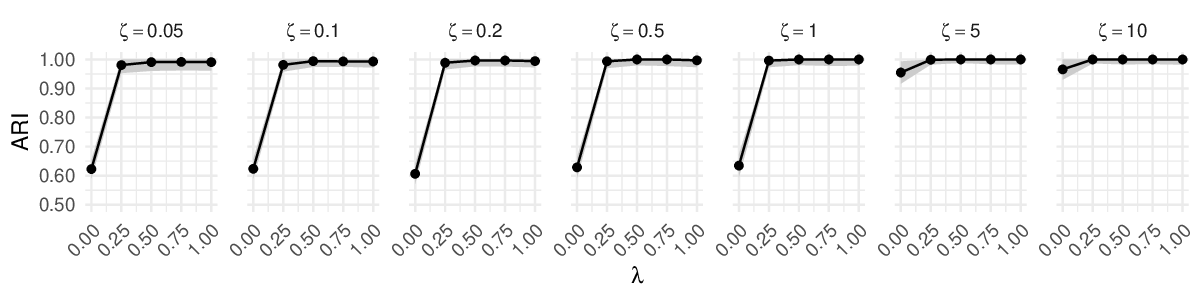}
\includegraphics[width=\linewidth]{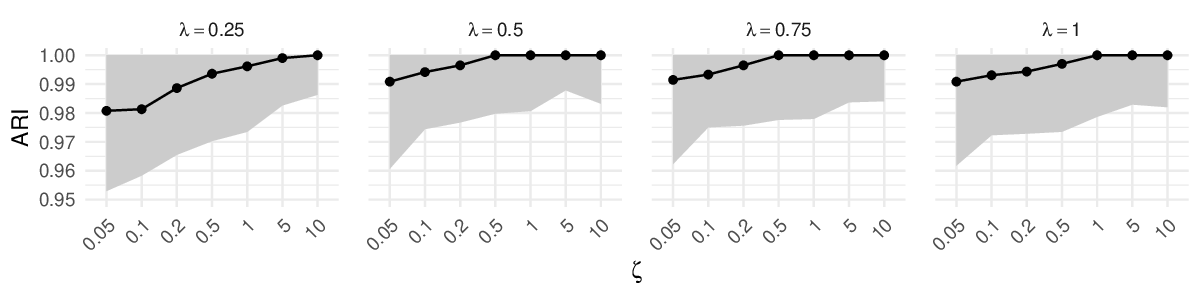}
\label{fig:fwjm_C_out_K3}
\end{subfigure}
\begin{subfigure}{.48\linewidth}
\centering
\caption{$K=3$, $\alpha=5\%$}
\includegraphics[width=\linewidth]{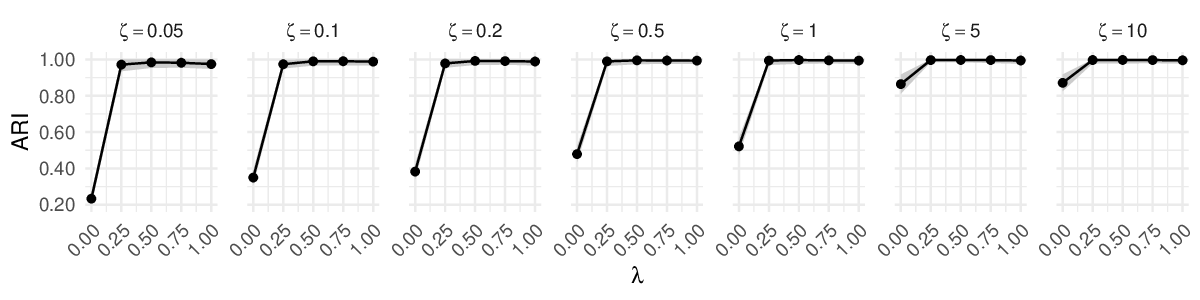}
\includegraphics[width=\linewidth]{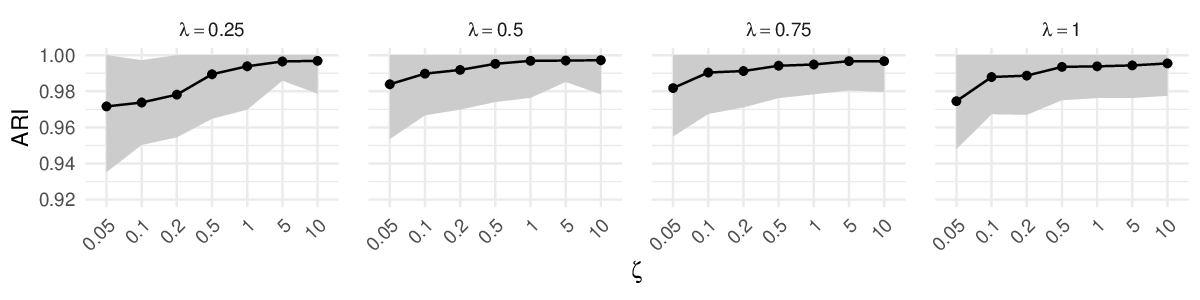}
\label{fig:fwjm_C_out_K3}
\end{subfigure}
\hfill
\begin{subfigure}{.48\linewidth}
\centering
\caption{$K=4$, $\alpha=0\%$}
\includegraphics[width=\linewidth]{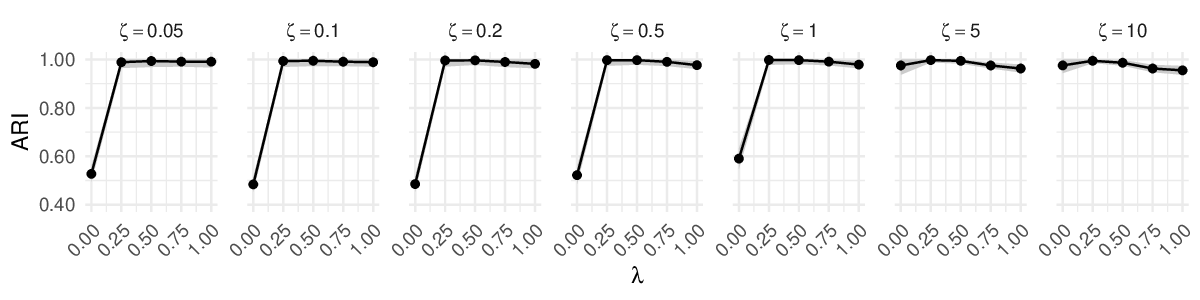}
\includegraphics[width=\linewidth]{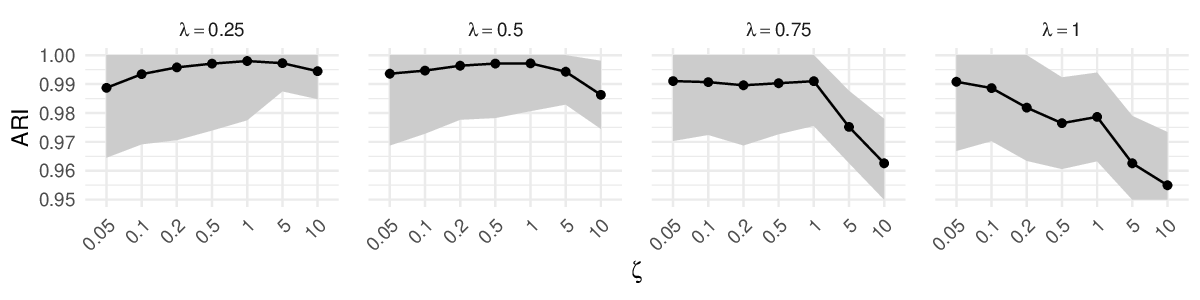}
\label{fig:fwjm_C_out_K3}
\end{subfigure}
\begin{subfigure}{.48\linewidth}
\centering
\caption{$K=4$, $\alpha=5\%$}
\includegraphics[width=\linewidth]{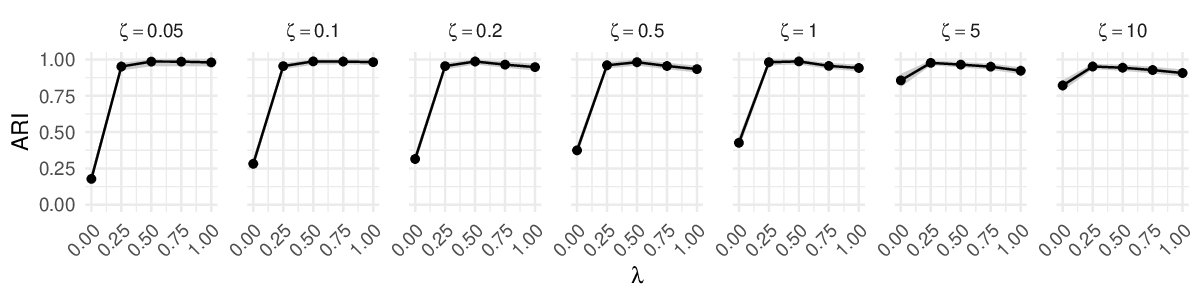}
\includegraphics[width=\linewidth]{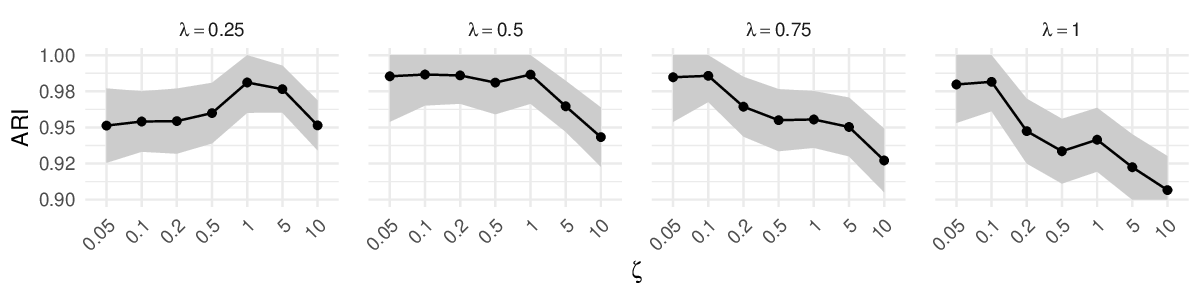}
\label{fig:fwjm_C_out_K3}
\end{subfigure}

\caption{
  Sensitivity analysis for the feature-weighted jump model (FWJM) under Scenario C, with $0\%$ and $5\%$ contamination, for $K=2, K=3$ and $K=4$ latent states. Each panel reports the median adjusted Rand index (ARI) as a function of the tuning parameters $\lambda$ and $\zeta$.
  Grey shaded areas denote confidence intervals.
}
\label{fig:sens_C}
\end{figure}

Figure~\ref{fig:sens_C} reports the sensitivity analysis for Scenario C, where clustering performance is generally high and robust to the choice of both hyperparameters.
%

\begin{figure}[!htbp] 
\centering
\begin{subfigure}{.48\linewidth}
\centering
\caption{$K=2$, $\alpha=0\%$}
\includegraphics[width=\linewidth]{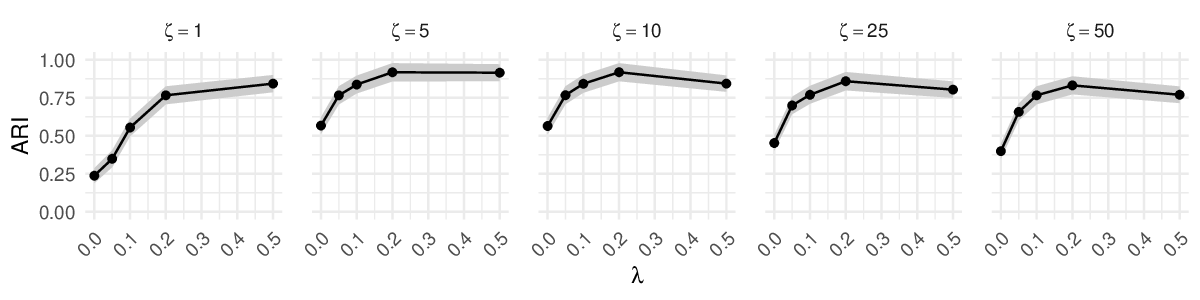}
\includegraphics[width=\linewidth]{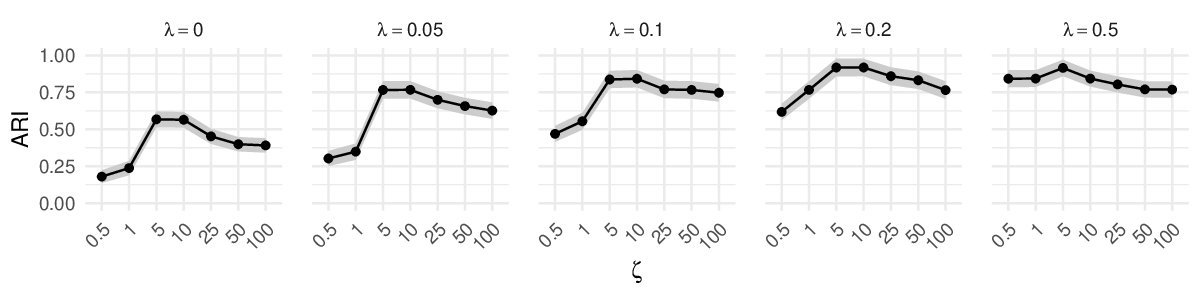}
\label{fig:fwjm_D_out_K3}
\end{subfigure}
\begin{subfigure}{.48\linewidth}
\centering
\caption{$K=2$, $\alpha=5\%$}
\includegraphics[width=\linewidth]{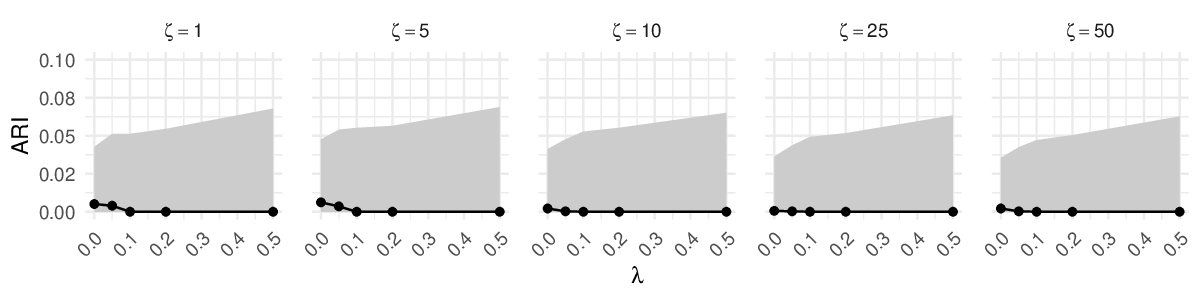}
\includegraphics[width=\linewidth]{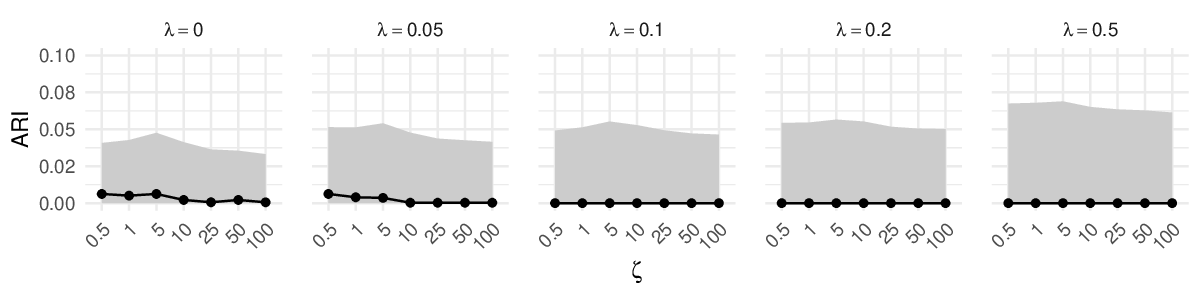}
\label{fig:fwjm_D_out_K3}
\end{subfigure}
\hfill
\begin{subfigure}{.48\linewidth}
\centering
\caption{$K=3$, $\alpha=0\%$}
\includegraphics[width=\linewidth]{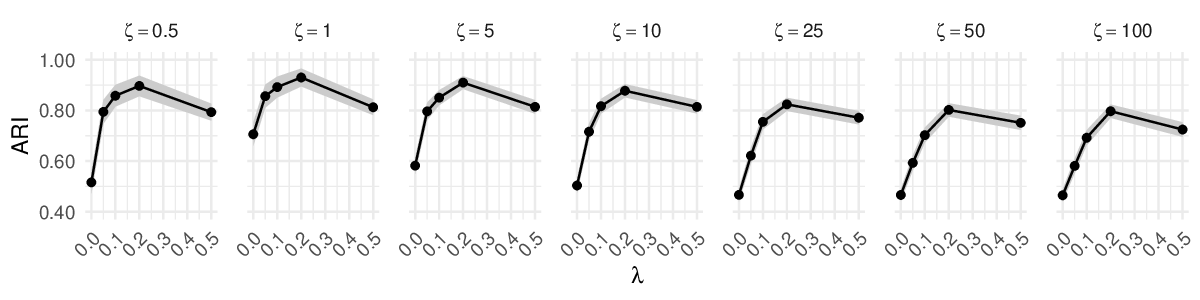}
\includegraphics[width=\linewidth]{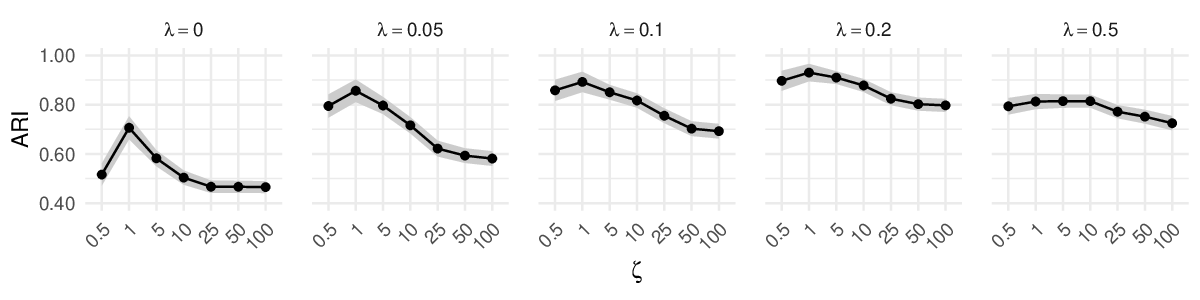}
\label{fig:fwjm_D_out_K3}
\end{subfigure}
\begin{subfigure}{.48\linewidth}
\centering
\caption{$K=3$, $\alpha=5\%$}
\includegraphics[width=\linewidth]{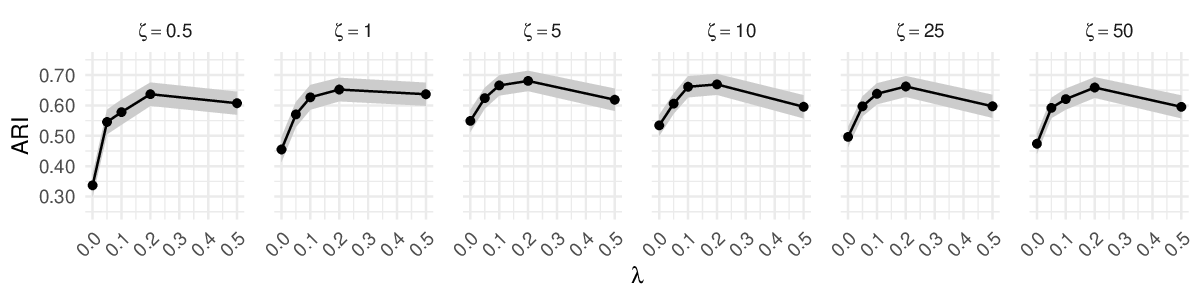}
\includegraphics[width=\linewidth]{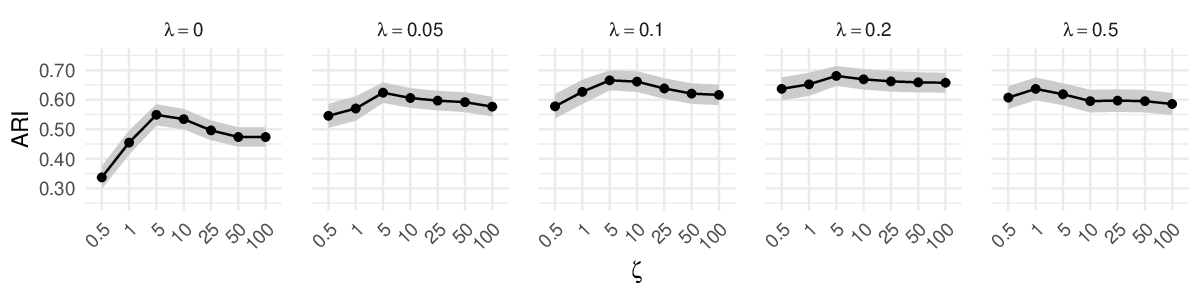}
\label{fig:fwjm_D_out_K3}
\end{subfigure}
\hfill
\begin{subfigure}{.48\linewidth}
\centering
\caption{$K=4$, $\alpha=0\%$}
\includegraphics[width=\linewidth]{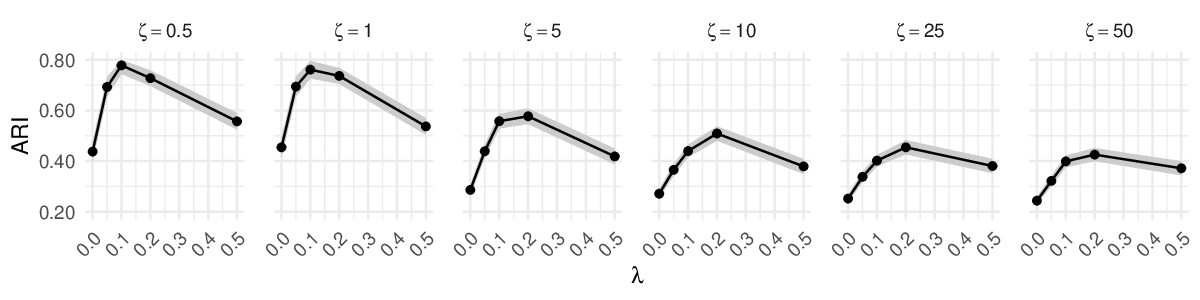}
\includegraphics[width=\linewidth]{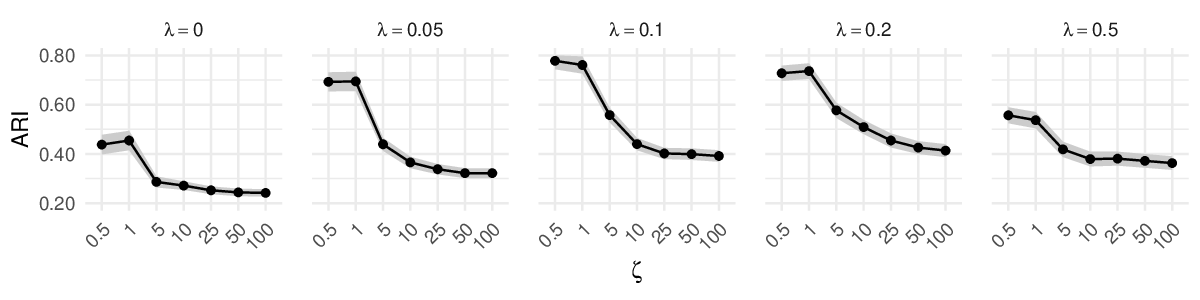}
\label{fig:fwjm_D_out_K3}
\end{subfigure}
\begin{subfigure}{.48\linewidth}
\centering
\caption{$K=4$, $\alpha=5\%$}
\includegraphics[width=\linewidth]{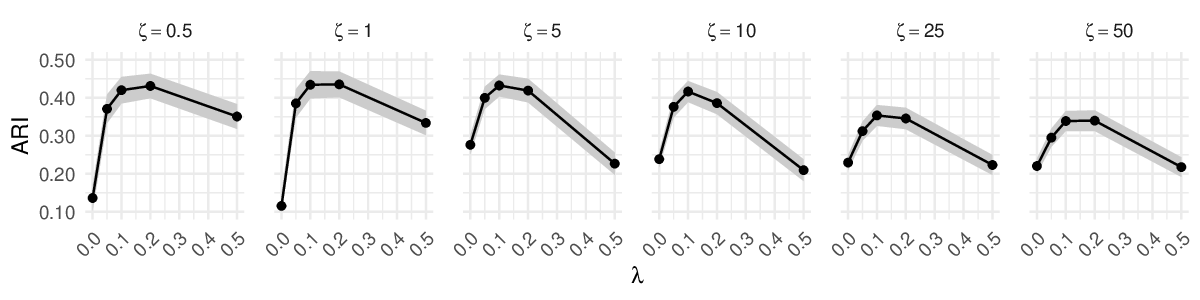}
\includegraphics[width=\linewidth]{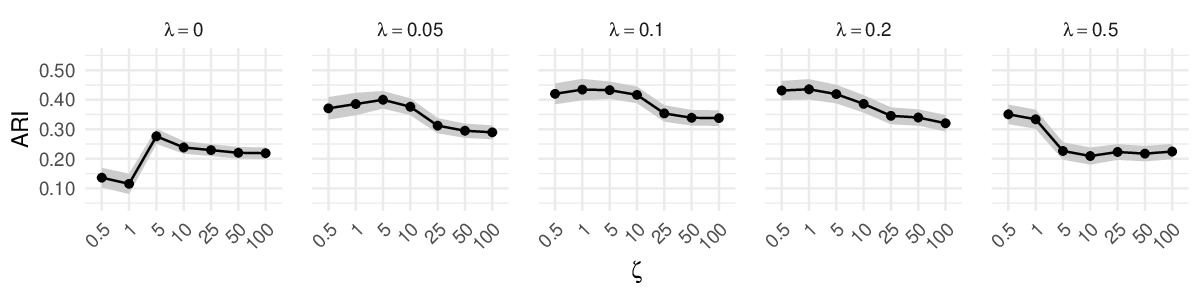}
\label{fig:fwjm_D_out_K3}
\end{subfigure}

\caption{
  Sensitivity analysis for the feature-weighted jump model (FWJM) under Scenario D, with $0\%$ and $5\%$ contamination, for $K=2, K=3$ and $K=4$ latent states. Each panel reports the median adjusted Rand index (ARI) as a function of the tuning parameters $\lambda$ and $\zeta$.
  Grey shaded areas denote confidence intervals.
}
\label{fig:sens_D}
\end{figure}

Figure~\ref{fig:sens_D} reports the sensitivity analysis for Scenario D, where satisfactory performance is generally achieved only in the absence of outliers.
Similarly to Scenario B, the temporal regularization parameter $\lambda$ must remain relatively small, typically below approximately $0.2$.
Likewise, smaller values of $\zeta$, generally between $1$ and $5$, provide the best result.

\clearpage
\section{Derivation of the feature weights update step}
\label{sec:deriv}

\label{subsec:proof_ordinary}
Considering the objective function
\begin{equation}
    \label{eq:FWJM}
    f(\boldsymbol{s},\boldsymbol{W},\boldsymbol{m})=  
   \sum_t \sum_p w_{s_t,p}d_{tm_{s_t},p}+
   \zeta\sum_k\sum_p w_{kp}\log w_{kp}+\lambda\sum_{t=1}^{T-1} \mathbb{I}(s_t\neq s_{t+1}), 
\end{equation}
We should solve 
$$\min_{\boldsymbol W}\;
\sum_p \sum_k w_{kp} \sum_{t:s_t=k}d_{t m_{s_t},p}
+\zeta\sum_k\sum_p w_{kp}\log w_{kp}
\quad
\text{s.t. }
\sum_{p=1}^P w_{kp}=1,\;\; w_{kp}\ge 0.$$
Define 
$$
S_{kp}
=
\sum_{t:s_t=k} d_{t m_k,p},
$$
then
$$
\min_{ w_{kp}}\;
\sum_{p=1}^P w_{kp}S_{kp}
+\zeta\sum_{p=1}^P w_{kp}\log w_{kp}
\quad
\text{s.t. }
\sum_{p=1}^P w_{kp}=1.
$$
Defining the Lagrangian as
$$
\mathcal L_k
=
\sum_{p=1}^P w_{kp}S_{kp}
+\zeta\sum_{p=1}^P w_{kp}\log w_{kp}
+\mu_k\left(\sum_{p=1}^P w_{kp}-1\right),
$$
and setting its first derivative with respect to $w_{kp}$ equal to zero, we obtain
\begin{equation}
\label{eq:wkp_lagr}
    w_{kp}=
\exp\left(-\frac{S_{kp}}{\zeta}\right)
\exp\left(-1-\frac{\mu_k}{\zeta}\right).
\end{equation}
By imposing the constraint
$
\sum_{p=1}^P w_{kp}
=
1
$,
we obtain that
$$
\mu_k
=
\zeta
\left[
\log\left(
\sum_{p=1}^P
\exp\!\left(-\frac{S_{kp}}{\zeta}\right)
\right)
-1
\right],
$$
and substituting in Eq. \eqref{eq:wkp_lagr}, we finally obtain
$$w_{kp}=\exp\left(-\frac{S_{kp}}{\zeta}\right)
\Bigg/\sum_{p=1}^P
\exp\!\left(-\frac{S_{kp}}{\zeta}\right).
$$

\end{document}